\documentclass[10pt, a4paper, twocolumn]{article}

\usepackage[dvips]{graphicx}
\usepackage{amsmath}
\usepackage{bm}
\usepackage{amsfonts}
\usepackage{url}
\usepackage{import}
\usepackage{caption}
\usepackage[subrefformat=parens]{subcaption}
\usepackage{times}

\title{Point Cloud Registration Based on Consistency Evaluation of Rigid
Transformation in Parameter Space}
\author{ Masaki Yoshii, Ikuko Shimizu}
\date{\empty}

\providecommand{\tightlist}{%
  \setlength{\itemsep}{0pt}\setlength{\parskip}{0pt}}

\newcommand{\argmax}{\mathop{\rm argmax}\limits}

\begin{document}

\maketitle

\begin{abstract}
We can use a method called registration to integrate some point clouds that represent the shape of the real world.
In this paper, we propose highly accurate and stable registration method.
Our method detects keypoints from point clouds and generates triplets using multiple descriptors.
Furthermore, our method evaluates the consistency of rigid transformation parameters of each triplet with histograms 
and obtains the rigid transformation between the point clouds.
In the experiment of this paper, our method had minimul errors and no major failures. 
As a result, we obtained sufficiently accurate and stable registration results compared to the comparative methods.
\end{abstract}

\section{Introduction}

We can obtain point clouds that represent the 3-dimensional shape of the
real world by using depth sensors. However the sensors, at one time,
can obtain only a partial shape within the measurement range. Therefore,
we have a demand for integrating measured objects (point clouds) in
order to restore them on a computer. To do that, there is a method
called registration.

Registration is a method of calculating the relationship between local
coordinates of point clouds. To calculate this relationship, in
ragistration, point correspondences are created between the point
clouds. The points associated with the correct correspondence represent
the same position of the real object. By obtaining a plurality of
correct correspondences, it is possible to calculate the rigid
transformation between the local coordinates of points clouds. Some
registration method can calculate non-rigid transformation.

Registration is becoming more and more important these days, but the
size of the acquired point cloud is also increasing as the performance
of hardware becomes higher. Most of the (fine) registration methods so
far have used all points in the point cloud. However, such methods
becomes difficult to calculate within a realistic time when the size of
point cloud becomes large.

Therefore, to perform the registration between large-size point clouds,
the points required for the registration, that is the keypoints, are
extracted from the point cloud and the registration is performed using
only the keypoints. This makes it possible to speed up the calculation.


Moreover, the robustness of descriptors can be improved by using keypoints.
For the descriptor of the type for which the valus of the vector are calculated using the local point cloud,
 the shape of the local point cloud must be characteristic.
If inappropriate points are used, a plurality of similar descriptors are obtained, and correct correspondences may not be obtained using those descriptors.
Therefore, it is important to detect keypoints when using descriptors.

The purpose of this paper is to propose a global registration method with high accuracy and stability at a practical speed.

The essential of our method is, first, to use keypoints. 
By using keypoints, we can obtain the robustness of the descriptors
 while improving the speed by reducing the points.
The second essential is to search a region with the correct triplets
 by expanding one-to-one correspondences to triplets
 and performing statistical evaluation of rigid transformation parameters obtained from each triplet using the histograms.
In detail, the correct triplet is a triplet with a small error from the ground truth of the rigid transformation.
At this time, since the error between the correct triplets is also small, they are concentrated around the ground truth.
On the other hand, the triplets with a large error from the ground truth - the wrong triplets - are dispersed in the region away from the ground truth, so there is basically no crowd of the wrong triplets.
Therefore, by searching for the dense region of the distribution of rigid transformation parameters, we can find the region where correct triplets are concentrated, that is, the region as close as possible to the ground truth.
By doing so, we can directly obtain the globally-optimum result instead of iterative calculation while removing the wrong correspondences.

Section 2 introduces the related registration method, section 3 describes the details of our method,
section 4 conducts experiments and evaluations on accuracy and speed between our method and the comparative methods,
and finally Section 5 concludes the paper.

\section{Related work}

There are various methods for registration, and the most famous of them
is Iterative Closest Point (ICP) proposed by P. J.
Besl\cite{ICPpoint2point1992} in 1992. At the same time, Y.
Chen\cite{ICPpoint2plane1991} proposed a similar method. There are many
derivative methods such as color ICP proposed by A. E.
Johmson\cite{colorICP1999} in 1999, ICPIF proposed by G. C.
Sharp\cite{ICPIF2000} in 2000, EM-ICP proposed by S.
Granger\cite{EM-ICP2002} in 2002, LM-ICP proposed by A.
Fitzgibbon\cite{LM-ICP2003} in 2003 and so on. In the most basic ICP,
when two point clouds are given, the transformation is iteratively
performed so that the mean square error of the distance between each
point and the nearest point in another point cloud is minimized. 
ICP and its derivatives are generally methods of obtaining a local optimal
solution using the all points, but in recent years, a method called Globally Optimal ICP
(Go-ICP) has been proposed by J. Yang\cite{Go-ICP2013, Go-ICP2016}.
Go-ICP makes it possible to obtain a global optimal solution by using
the ICP and the Branch-and-Bound (BnB) method.

There are other methods that use Gaussian mixture model (GMM). These
methods are the derivatives of the one proposed by B.
Jian\cite{RegGMM2005} in 2005 based on the framework proposed by G.
Chui\cite{RegGMMframework2000} in 2000.
These methods are the global registration methods using all points,
which can obtain more robustness to noise than ICP 
by treating the point cloud as a GMM.
Also, depending on the method, non-rigid transformation can be obtained.
As derivative methods, there are Coherent Point Drift (CPD) proposed by A. Myronenko\cite{GMM_CPD2010} in
2010, GMMerge proposed by D. Campbell\cite{GMMerge2015} in 2015,
adaptive GMM proposed by C. Pu\cite{adaptiveGMM2017} in 2017 and so on.

Furthermore, as registration methods other than the above
classification, there are 4-Points Congruent Sets (4PCS) proposed by D.
Aiger\cite{4PCS2008} in 2008, Fast Global Registration (FGR) proposed by
Q.-Y. Zhou\cite{FGR2016} in 2016, Discriminative Optimization (DO)
proposed by J. Vongkulbhisal\cite{DO2017} in 2017 and so on. 
4PCS is a global registration method that selects a combination of four points from all points;
the four points are on the same plane that are invariant to the affine transformation.
The method can be used even when the overlap between point clouds is small compared to ICP.
and there is also a derivative method called Super 4-Points Congruent Sets
(Super4PCS)\cite{SUPER4PCS2014} that speeds up. 
FGR is a global registration method that minimize the function of rigid transformation using Geman-McClure Estimator as a penalty function. 
That is a high-speed method that combines simple methods, and unlike ICP, it does not perform iterative calculation using all points.
DO is a global registration method using all points that based on SSU (Supervised Sequential Update),
which is a method of detecting and tracking facial organs.
By learning from the data, this method does not fall into a local minimum unlike ICP.
Inverse Composition Discriminative Optimization (ICDO)\cite{ICDO2018}, which speeded up and generalized DO, exists as a derivative method.

\section{Method}

In this chapter, we explain the process to generate triplets for the
evaluation in the experiment.

Suppose two point clouds \(X\) and
\(Y\)\(\subset \mathbb{R}^{3 \times 1}\) are given. At this time, using
the points \({\bf x} \in X\) and \({\bf y} \in Y\) in each point cloud
\(X\), \(Y\),\\
triplets are generated by the following flow.

\begin{enumerate}
\def\labelenumi{\arabic{enumi}.}
\tightlist
\item
  Calculate the normals \({\bf n}_x\), \({\bf n}_y\) for the points
  \({\bf x} \in X\), \({\bf y} \in Y\).
\item
  Find the sets of keypoints \(P_x\) and \(P_y\) for each point cloud
  \(X\), \(Y\).
\item
  Calculate FPFH descriptors\cite{FPFH2009_1, FPFH2009_2}
  \(f({\bf p}_x)\) and \(f({\bf p}_y)\) for each keypoint
  \({\bf p}_x \in P_x\), \({\bf p}_y \in P_y\).
\item
  Generate set of one-to-one correspondences \(C_1\) from keypoint sets
  \(P_x\), \(P_y\). At this time, FPFH descriptors \(f({\bf p}_x)\),
  \(f({\bf p}_y)\) are used. AApproximate curvatures
  \(\tilde{\sigma}({\bf x})\), \(\tilde{\sigma}({\bf y})\) are also
  used.
\item
  Generate set of triplets \(C_3\) from set of one-to-one
  correspondences \(C_1\).
\item
  Estimate a rigid transformation from each tripret \(c_3 \in C_3\), and
  calculate the rotation vector \({\bf r}\) and the translation vector
  \({\bf t}\). Obtain the mode from each of the rotation vector
  \({\bf r}\) and the translation vector \({\bf t}\). The mode is used
  as the estimated rigid transformation between point clouds.
\end{enumerate}

In this paper, we do not specify the keypoints detection method, but will propose in the original of this paper. 
The details of each stage are as follows.

\subsection{Calculation of normals}

The method of principal component analysis (PCA) is used to calculate
the normals. First, the local point cloud \(Z \subset X\) is extracted
using kd-tree for the query point
\({\bf x} \in X \subset \mathbb{R}^{3 \times 1}\). Next, the covariance
matrix \(\Sigma\) of \(Z\) centered on \({\bf x}\) is calculated by the
following formula.

\begin{equation}
\Sigma = \frac{1}{|Z|}\sum_{{\bf z} \in Z}({\bf z} - {\bf x})({\bf z} - {\bf x})^{\rm T}
\end{equation}

Furthermore, PCA is performed on \(\Sigma\) to obtain eigenvalues
\(\lambda_1 \geq \lambda_2 \geq \lambda_3\) and corresponding
eigenvectors \({\bf u}_1\), \({\bf u}_2\), \({\bf u}_3\). Eigenvector
\({\bf u}_3\) corresponding to the minimum eigenvalue \(\lambda_3\)
becomes the normal \({\bf n}_x\) of \({\bf x}\). The sign is corrected
in consideration of the measurement direction and the like because the
sign of this normal is indefinite.




By this method, we obtain the normals for all points in \(X\). We also do the same calculation for \(Y\).

Here, for each point cloud \(X\), \(Y\), get the medians of distribution
of distances between the points \({\bf x}\), \({\bf y}\) and their
nearest points. These medians are named \({\rm med} D_X\) and
\({\rm med} D_Y\), and record
\({\rm med}D = ({\rm med} D_X + {\rm med} D_Y) / 2\). This
\({\rm med}D\) will be used for some parameter calculations.

\subsection{Calculation of keypoints and descriptors}

When the calculation of the normals is completed, the keypoints \(P_x\) and \(P_y\)
are extracted from the point clouds \(X\) and \(Y\).

When \(P_x\) and \(P_y\) are obtained, the descriptors are calculated
next. In this paper, the FPFH descriptors\cite{FPFH2009_1, FPFH2009_2}
\(f({\bf p}_x)\) and \(f({\bf p}_y)\) are used. FPFH descriptors are
calculated for all keypoints in \(P_x\), \(P_y\).

\subsection{Calculation of correspondences}

When two keypoint sets \(P_x\), \(P_y\) are obtained, a set of
one-to-one correspondences \(C_1\) is generated using those.

To generate \(C_1\), FPFH descriptors \(f({\bf p}_x)\), \(f({\bf p}_y)\) are used. 
In some cases, other values may be used additionally.
For each keypoint \({\bf p}_x \in P_x\), using \(f({\bf p}_x)\) as a query, the local
point cloud \({\rm knn}(f({\bf p}_x)) \subset P_y\) is searched from a
kd-tree composed of \(f({\bf p}_y)\). \(C_1\) satisfies the following
equation.

\begin{equation}
\begin{aligned}
C_1' =& \{ ({\bf p}_x,~{\bf p}_y) \mid \forall {\bf p}_x \in P_x, \\
& ~{\bf p}_y \in {\rm knn}(f({\bf p}_x)) \\
& \land ~[{\rm ~other~restrictions}\dots~] \}
\end{aligned}
\end{equation}

\subsection{Calculation of triplets}

When the set of one-to-one correspondences \(C_1\) is obtained, next
calculate a set of triplets \(C_3\). However, if \(C_3\) is
straightforwardly calculated, the number will be enormous. Therefore, by
performing the calculation according to the procedure described below,
our method obtains a fixed number of triplets \(C_3\) having the desired
properties.

First, for each one-to-one correspondence
\(c_1^{(i)} = ({\bf p}_x^{(i)}, {\bf p}_y^{(i)}) \in C_1\), the
consistency with other correspondence \(c_1^{(j)}\) is evaluated, and
the total of those value \(\tilde{\pi}_i\) is taken as the reliability
of \(c_1^{(i)}\). The reliability \(\tilde{\pi}_i\) is calculated by the
following formula, where \(h\) is a parameter that determines the range
and is set to \(h = h_r {\rm med}D\) (\(h_r\) is the size with
\({\rm med}D\) as the unit).

\begin{equation}
\begin{aligned}
\tilde{\pi}_i =& \sum_{c_1^{(j)} \in C_1} \frac{1}{1 + h^{-2}(d_x^2 - d_y^2)^2} \\
& d_x = || {\bf p}_x^{(i)} - {\bf p}_x^{(j)} || \\
& d_y = || {\bf p}_y^{(i)} - {\bf p}_y^{(j)} ||
\end{aligned}
\end{equation}

Further, since \(\tilde{\pi}_i\) does not require accuracy as much as
that, the speed is increased by dividing \(C_1\). In detail, \(C_1\) is
divided into some subsets \(C_1^{k}\), and \(\tilde{\pi}_i\) is
calculated between the correspondences in each subset
\(c_1^{(i_k)} \in C_1^{k}\). At this time, \(\tilde{\pi}_i\) is
multiplied by the coefficient \(|C_1|/|C_1^{k}|\) to correct the
difference in the number of elements between the subsets.

When the reliabilities are obtained, sort \(C_1\) in descending order
using these values and generate the triplets \(C_3\) while calculating
the directed graph between the \(C_1\), where each ordered
correspondence \(c_1^{(i)} \in C_1\) is a node of graph.

The calculation of the branches of the graph and the generation of
\(C_3\) are alternately performed by the following procedure.

\begin{enumerate}
\def\labelenumi{\arabic{enumi}.}
\tightlist
\item
  \(i := 1\)
\item
  For \(c_1^{(i)}\), perform a similarity of the PPF descriptor with all
  \(c_1^{(j)} (j < i)\), and connect the branches with \(c_1^{(i)}\) as
  the starting node side when they are similar.
\item
  Follow the branch only twice from \(c_1^{(i)}\). When the search is
  performed in the order of
  \(c_1^{(i)} \rightarrow c_1^{(j)} \rightarrow c_1^{(k)}~(k <j <i)\),
  the three one-to-one correspondence
  \((c_1^{(i)},~c_1^{(j)},~c_1^{(k)})\) obtained at this time are
  candidates for a triplet \(c_3\). Similarity evaluation of PPF
  descriptors\cite{PPF2010} is performed between \(c_1^{(i)}\) and
  \(c_1^{(k)}\), and if they are similar, further angle evaluation is
  performed. Those that pass all evaluation are considered a triplet
  \(c_3\).
\item
  Repeat step 3 for all possible combinations.
\item
  If \(|C_3| \geq N_{C3\min}\), terminate the process. Also if
  \(i \geq s|C_1|\) is reached, the process is terminated.
\item
  Return to step 2 as \(i := i + 1\).
\end{enumerate}

Where \(N_{C3\min}\) is lower limit of \(|C_3|\) and \(s\) is a
parameter that determines the ratio to the number of \(C_1\).

Here, the similarity evaluation of two PPF descriptors
\((F_1,F_2,F_3,F_4)\) and \((G_1,G_2,G_3,G_4)\) is performed by the
following formula, where \(\vartheta_\gamma~(\gamma=1,2,3)\) are the
thresholds.

\begin{equation}
\begin{aligned}
\left( \vartheta_1 < \frac{F_1}{G_1} < \frac{1}{\vartheta_1} \right) \land (| F_2 - G_2 | < \vartheta_2) \\
 \land (| F_3 - G_3 | < \vartheta_2) \land (| F_4 - G_4 | < \vartheta_3)
\end{aligned}
\end{equation}

Since \(F_1\) and \(G_1\) are the distances between points, they are
evaluated by the ratio. The other values are angles, so they are
evaluated by the difference. Because \(F_4\) and \(G_4\) are the angles
between the normals of two points, they are less accurate than the other
values and different thresholds are used.

Also, the angle evaluation between the three one-to-one correspondences
\((c_1^{(i)},~c_1^{(j)},~c_1^{(k)})\) is performed to guarantee that the
three keypoints in same point cloud are not aligned on a straight line
when the rigid transformation is calculated from the triplet. The angle
evaluation is only performed on the minimum edge \(a < b, c\) (or
\(a^2 < b^2, c^2\)) of the triangle consisting of three keypoints At
this time, it is determined that other angles are also larger than that.
When the inequality
\(b^2 + c^2 - a^2 < 2 \sqrt{b^2 c^2} \cos \vartheta_{\triangle}\) is
satisfied, the angle evaluation is passed. Where
\(\vartheta_{\triangle}\) is the threshold. This evaluation is performed
on both triangles in \(X\) and \(Y\).

\subsection{Estimation of the rigid transformation between point clouds}

After generating the triplets \(C_3\), we estimate the rigid
transformation for each triplet \(c_3 \in C_3\). As the estimation
method, the one using singular value
decomposition\cite{EstimateTransformation1991} is used. The parameters
of the estimated rigid transformation are rotation angle
\(\theta \in \mathbb{R}\), rotation axis
\({\bm \alpha} \in \mathbb{R}^{3 \times 1}\) and translation
\({\bf t} \in \mathbb{R}^{3 \times 1}\), and the rotation vector
\({\bf r} = \theta {\bm \alpha}\) is calculated. A rotation vector and a
translation vector are found in all triplets, and sets of them are set
as a rotation vector set \(R\) and a translation vector set \(T\),
respectively.

First, for the set \(V \in \{R,~ T\}\), the covariance matrix

\begin{equation}
\Sigma_V = \frac{1}{|V|}\sum_{{\bf v} \in V}({\bf v} - \bar{\bf v})({\bf v} - \bar{\bf v})^{\rm T}
\end{equation} 
is calculated, and the principal component analysis (PCA) is
performed on \(\Sigma_V\). Where \(\bar{\bf v}\) is the average value of
\({\bf v} \in V\). The result of PCA is

\begin{equation}
{\rm PCA}(\Sigma_V) = U_V \Lambda_V U_V^{\rm T}
\end{equation}
. 
Further, the coordinate transformation of \({\bf v} \in V\) is performed by
\(\hat{\bf v} = U_V^{\rm T}{\bf v}\). Let \(\hat{V}_x~(x=1,2,3)\) be a
set of the elements of the vector \(\hat{\bf v}\) after the
transformation.

Next, create a histogram for \(\hat{V}_x\). The number of bins in the
histogram is \(B\), and the interval width \(h\) is determined using the
Freedman-Diaconis' choice\cite{Freedman1981}. The bin number \(b\) in
which \(\hat{v}_x \in \hat{V}_x\) is stored is determined by the
following formula, where \({\rm med} \hat{V}_x\) is the median of
\(\hat{V}_x\).

\begin{equation}
b = \left\lfloor \frac{\hat{v}_x}{h} - \frac{{\rm med} \hat{V}_x}{h} + \frac{B}{2} \right\rfloor
\end{equation}

If \(b\) is out of the range of \([0 \leq b \leq B-1]\), the vector
\(\hat{v}_x\) is not added to the histogram.

Next, when the set of elements contained in bin \(b\) of the histogram
is \(\hat{V}_x^{(b)}\), find \(\hat{v}_x^\dagger\) by the following
formula using the mode bin \(b^\ast = \argmax_b | \hat{V}_x^{(b)} |\)
and its \(\delta\) neighborhoods.

\begin{equation}
\hat{v}_x^\dagger = \frac{1}{|\hat{V}_x^{\ast}|}\sum_{\hat{v}_x^{\ast} \in \hat{V}_x^{\ast}} \hat{v}_x^{\ast}~~ \left( \hat{V}_x^{\ast} = \bigcup^{b^\ast + \delta}_{b = b^\ast - \delta} \hat{V}_x^{(b)} \right)
\end{equation}

Find this value in all cases (\(V\in\{R,~ T\}\), \(x=1,2,3\)) and the
resulting rotation vector is \(\hat{\bf r}^\dagger\) and the translation
vector be \(\hat{\bf t}^\dagger\). Finally, the coordinate of
\(\hat{\bf r}^\dagger\) and \(\hat{\bf t}^\dagger\) is restored by the
inverse transformation;
\({\bf r}^\dagger = U_R \hat{\bf r}^\dagger, ~~{\bf t}^\dagger = U_T \hat{\bf t}^\dagger\).
The calculated vectors \({\bf r}^\dagger\) and \({\bf t}^\dagger\) are
the estimated rigid transformation parameters between point clouds
\(X\), \(Y\). Note that \(\theta^\dagger = || {\bf r}^\dagger ||\),
\({\bm \alpha}^\dagger = {\bf r}^\dagger / \theta^\dagger\).

\section{Experimentation}

In this section, we conducted experiments and evaluations on accuracy and speed.
The following computer and software were used as the environment for this experiment.
We have not overclocked the CPU.

\begin{itemize}
 \setlength{\parskip}{0cm} 
 \setlength{\itemsep}{0cm} 
\item CPU: Intel Core i7 7700K
\item OS: Windows 10
\item Software: Microsoft Visual Studio 2015 Community Edition
\end{itemize}

\subsection{Used keypoints detection method}

In this experiment, we used approximate curvatures\cite{Rusu2013} 
$\tilde{\sigma}$ as the keypoints detection method. 
The following formula is used for the calculation.

\begin{equation}
\tilde{\sigma} = \frac{\lambda_3}{\lambda_1 + \lambda_2 + \lambda_3} \leq \frac{1}{3} 
\end{equation}

Where $\lambda_1$, $\lambda_2$ and $\lambda_3$ are eigenvalues of the covariance matrix of the local point cloud obtained during the normal calculation.

Several points are selected as the keypoints of $X$ in order from the point 
where the approximate curvature $\tilde{\sigma}({\bf x})$ becomes large, 
and the set of selected points is $P_x$.
Do the same for the point cloud $Y$ to obtain the keypoints $P_y$.

Furthermore, we added the following equation as a restriction when generating one-to-one correspondences.

\begin{equation}
| \tilde{\sigma}({\bf p}_x) - \tilde{\sigma}({\bf p}_y)| < \vartheta_{\tilde{\sigma} }
\end{equation}

Where $\tilde{\sigma}({\bf p}_x)$ and $\tilde{\sigma}({\bf p}_y)$ are the approximate curvature value corresponding to ${\bf p}_x$ and ${\bf p}_y$.
$\vartheta_{\tilde{\sigma} }$ is the threshold.

\subsection{Comparative methods}

In this experiment, we compared our method with other methods.
we used Fast Global Registration (FGR)\cite{FGR2016}, Coherent Point
Drift (CPD)\cite{GMM_CPD2010}, Globally-Optimal
ICP (Go-ICP)\cite{Go-ICP2013, Go-ICP2016} and Super 4-Points Congruent
Sets (Super4PCS)\cite{SUPER4PCS2014} as the comparative methods.

\subsection{Threshold settings}

The threshold settings in this experiment will be described in this
section. First, 1500 keypoints were generated, and the search radius of
the kd-tree of FPFH descriptors was set to \(10 \times {\rm med}D\). In
addition, the number of kd-tree searches in the one-to-one
correspondences generation \(K\) was \(15\) and the threshold of the
approximate curvature values used as constraints
\(\vartheta_{\tilde{\sigma}}\) was set to \(0.05\). Furthermore, in the
triplets generation, the range determination parameter of reliability
evaluation \(h_r\) was \(10\), the number of divisions was \(4\), PPF
descriptor thresholds \((\vartheta_1, \vartheta_2, \vartheta_3)\) were
\((0.95, 3.0 [{\rm degrees}], 5.0[{\rm degrees}])\), the triangle angle
threshold \(\vartheta_\triangle\) was \(20\) degrees, the thresholds of
the triplets' search termination were \(N_{C3 \min} = 150000\) and
\(s = 0.6\). Mode's neiborhoods parameter of histogram \(\delta\) was
\(1\).

\subsection{Used model data}

\begin{figure*}[t]
\centering
\begin{minipage}{.242\textwidth}
\includegraphics[width=\textwidth]{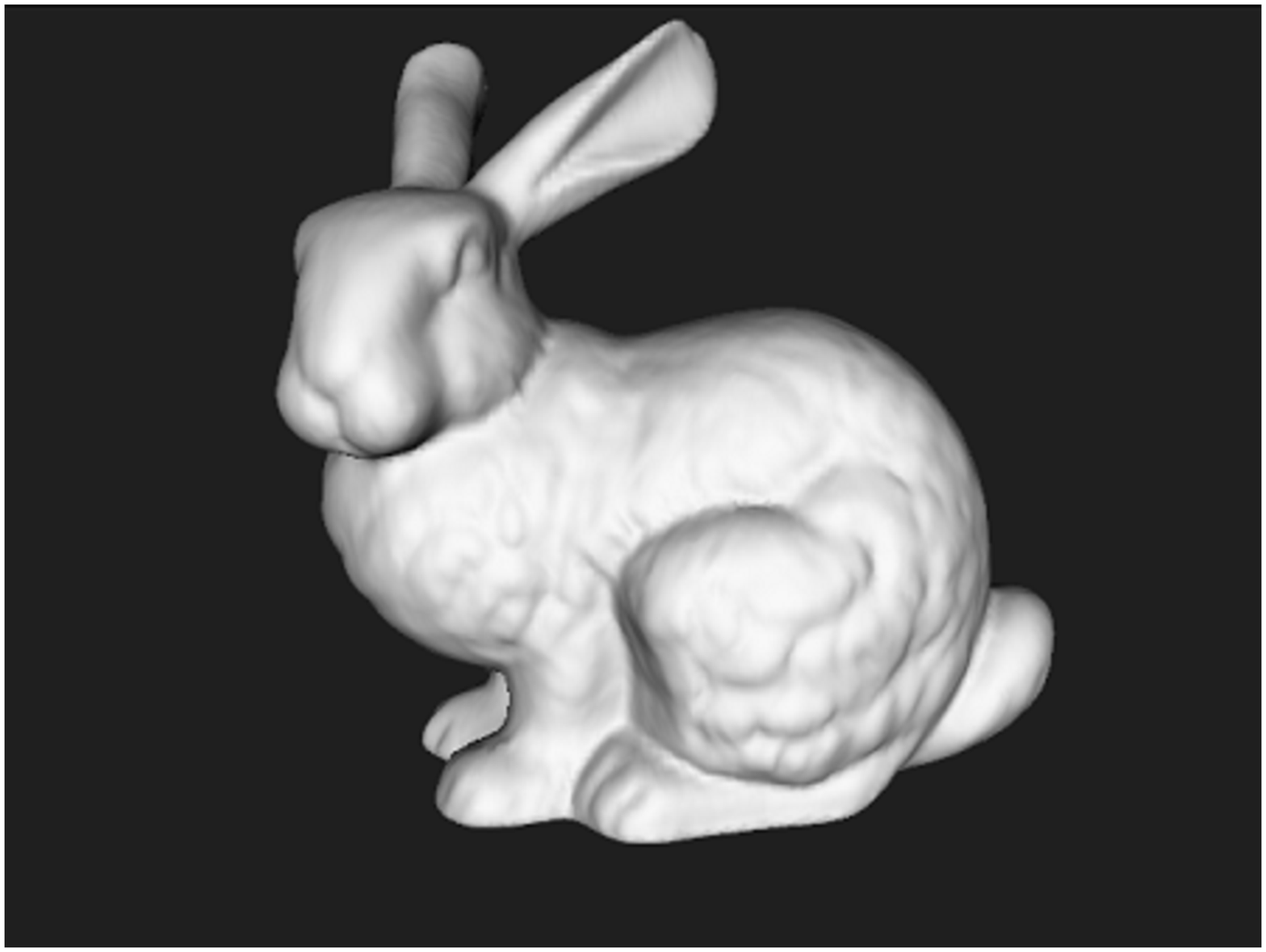}
\subcaption{bunny}
\label{Fig:StanfordModel_Bunny}
\end{minipage}
\begin{minipage}{.242\textwidth}
\includegraphics[width=\textwidth]{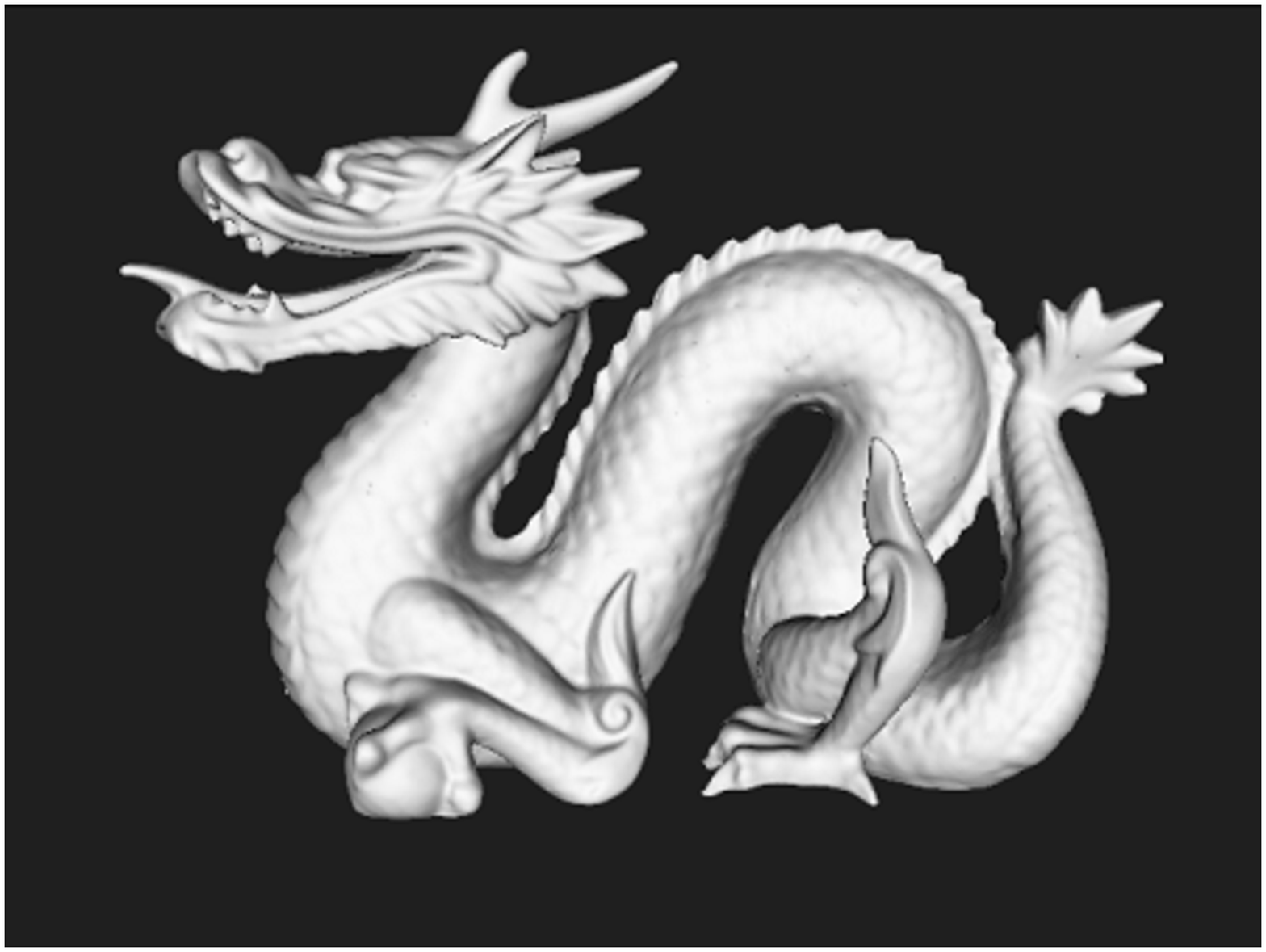}
\subcaption{dragon}
\label{Fig:StanfordModel_Dragon}
\end{minipage}
\begin{minipage}{.176\textwidth}
\includegraphics[width=\textwidth]{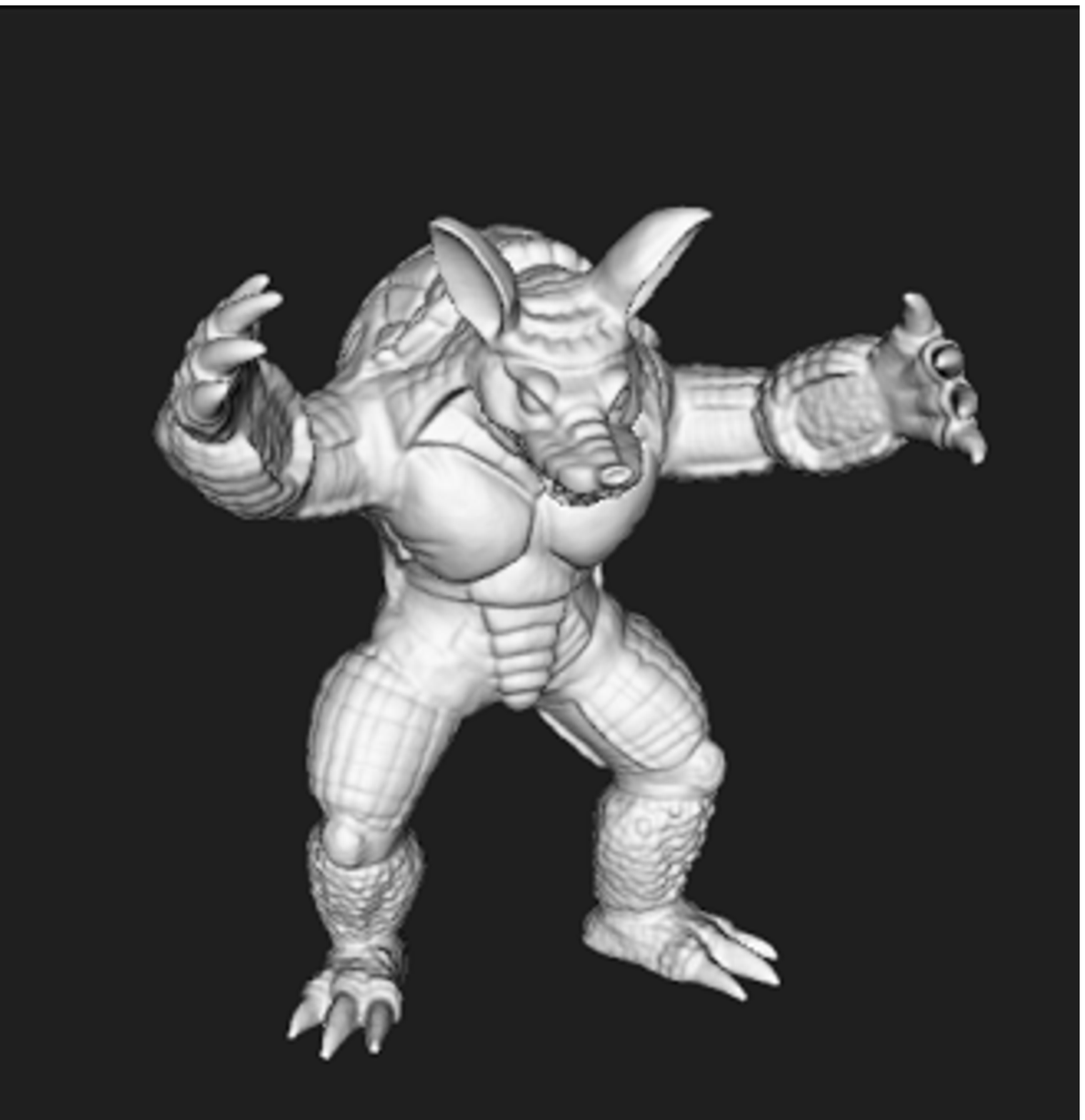}
\subcaption{armadillo}
\label{Fig:StanfordModel_Armadillo}
\end{minipage}
\begin{minipage}{.136\textwidth}
\includegraphics[width=\textwidth]{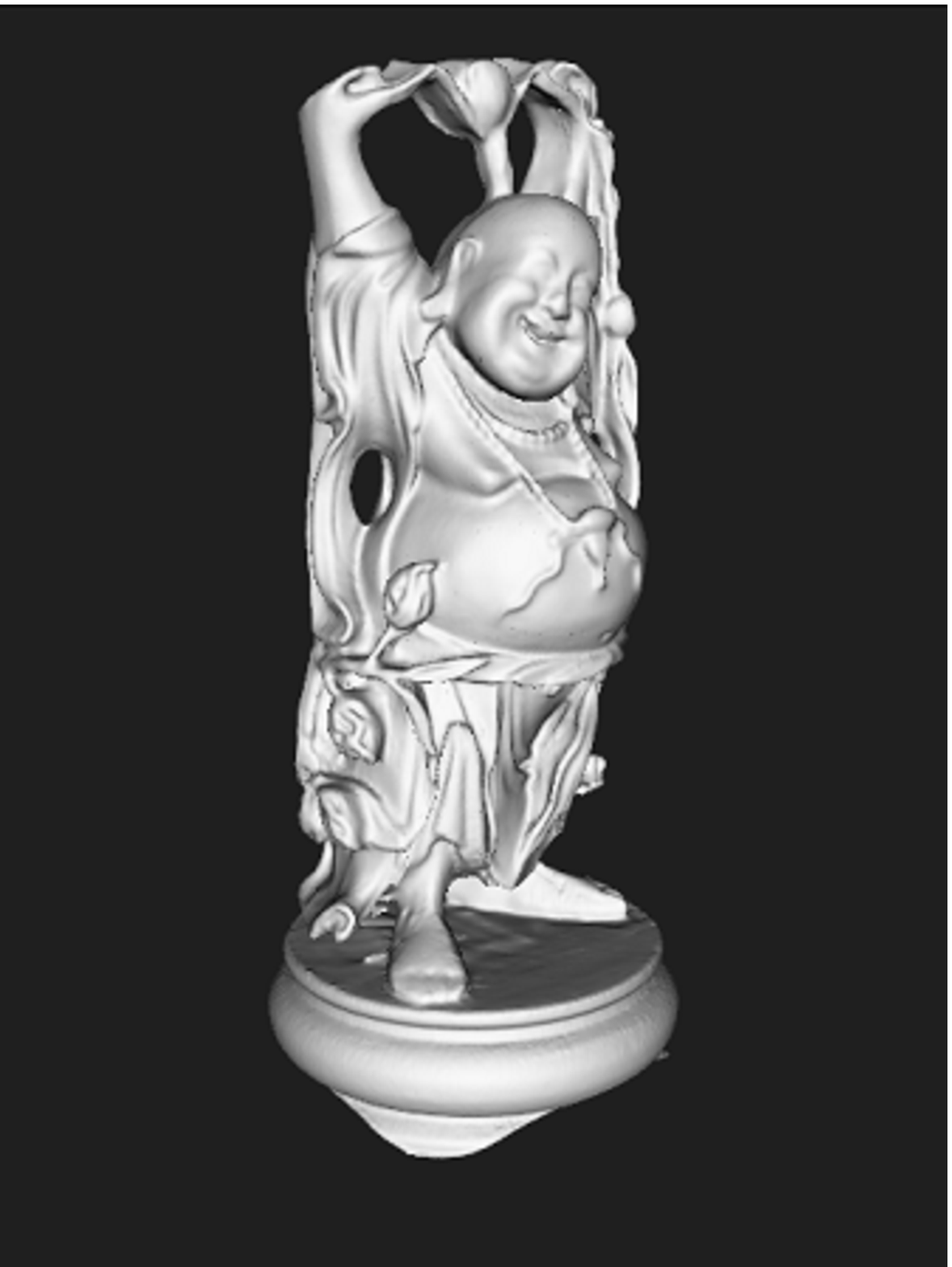}
\subcaption{buddha}
\label{Fig:StanfordModel_Buddha}
\end{minipage}
\begin{minipage}{.136\textwidth}
\includegraphics[width=\textwidth]{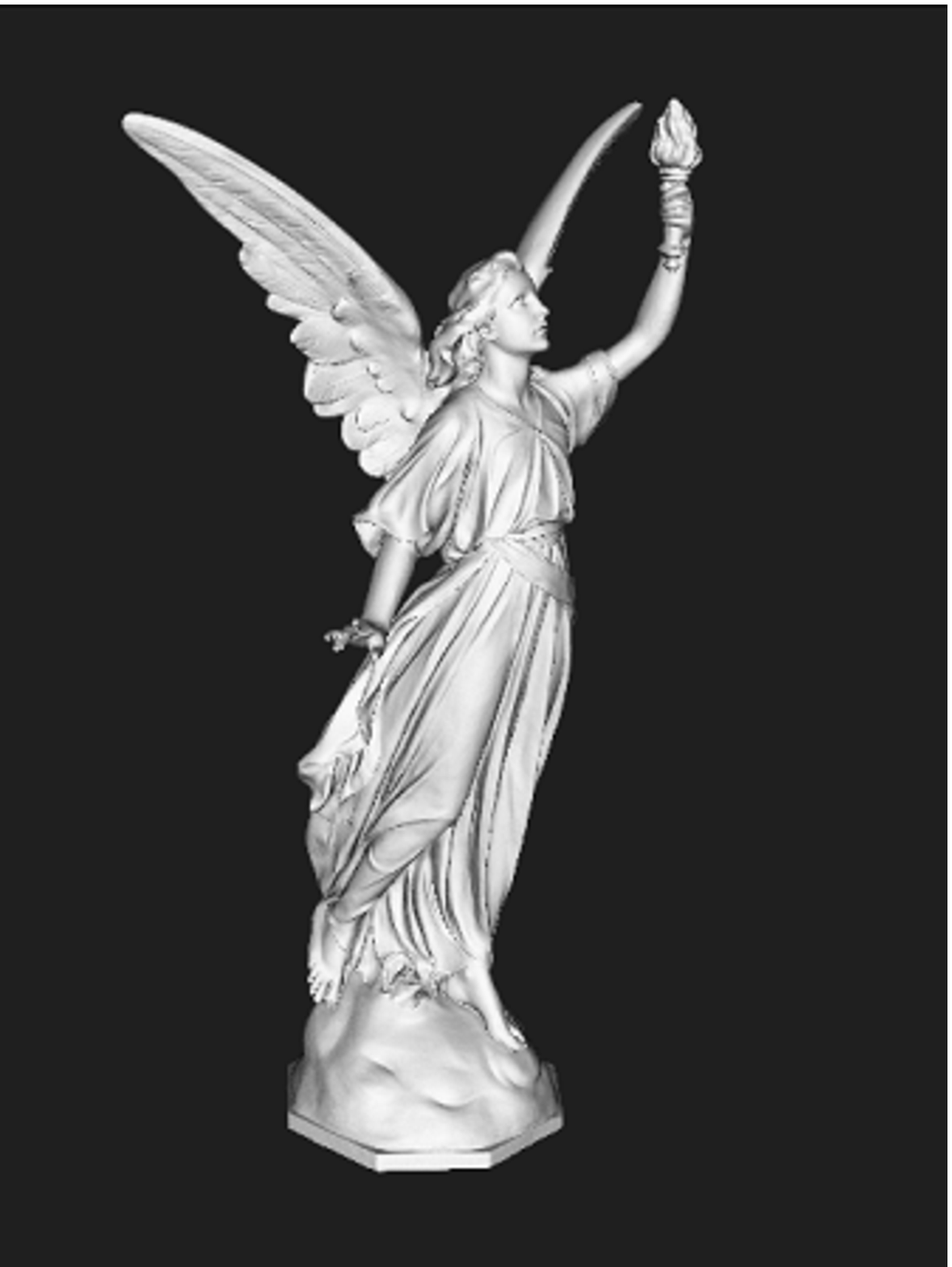}
\subcaption{lucy}
\label{Fig:StanfordModel_Lucy}
\end{minipage}

\caption{Stanford data}
\label{fig:stanfordModel}
\end{figure*}

\begin{figure*}[t]
\centering
\begin{minipage}{.19\textwidth}
\includegraphics[width=\textwidth]{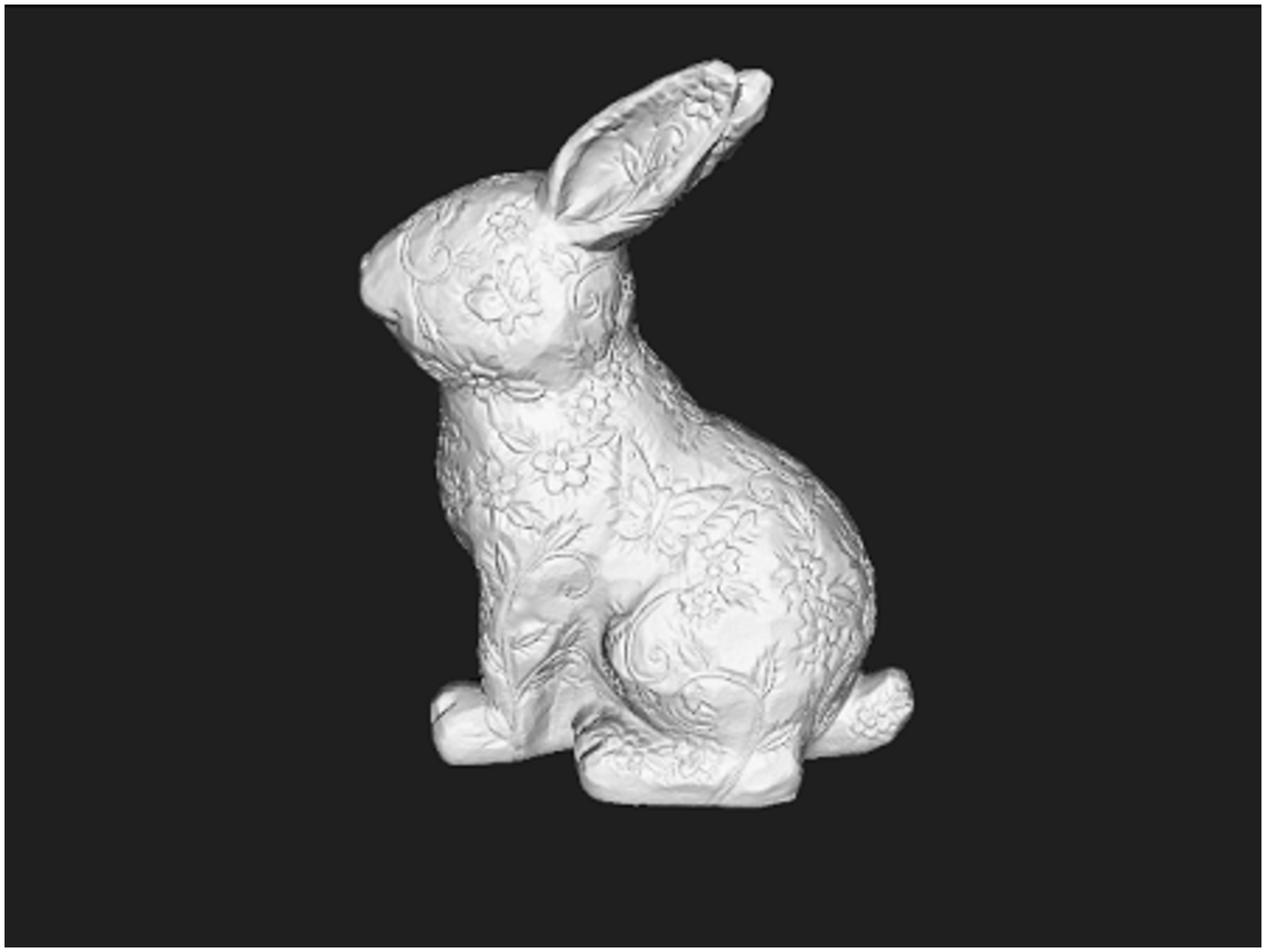}
\subcaption{lab bunny}
\label{Fig:LabModel_Bunny}
\end{minipage}
\begin{minipage}{.19\textwidth}
\includegraphics[width=\textwidth]{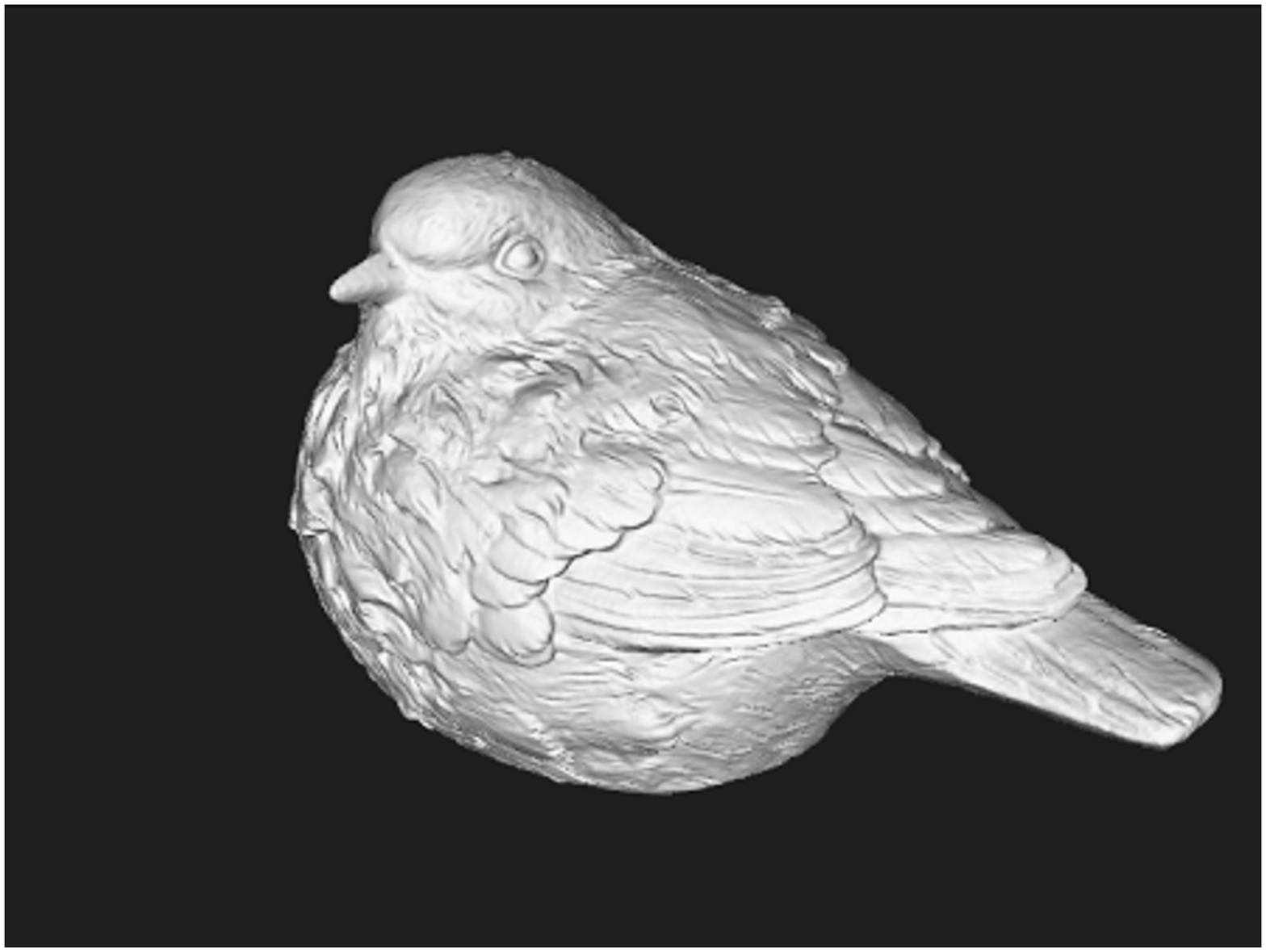}
\subcaption{bird}
\label{Fig:LabModel_Bird}
\end{minipage}
\begin{minipage}{.19\textwidth}
\includegraphics[width=\textwidth]{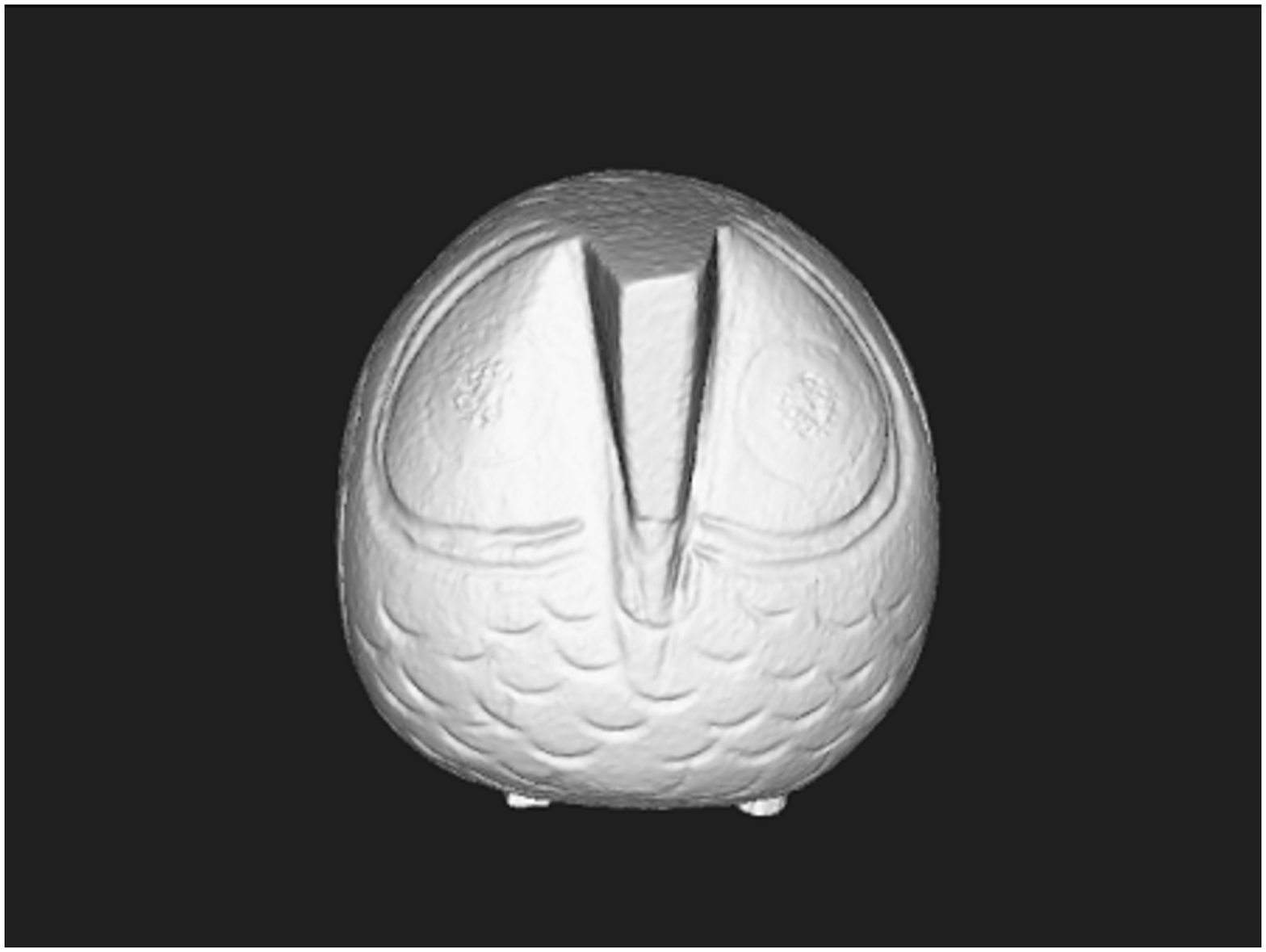}
\subcaption{owl}
\label{Fig:LabModel_Owl}
\end{minipage}
\begin{minipage}{.19\textwidth}
\includegraphics[width=\textwidth]{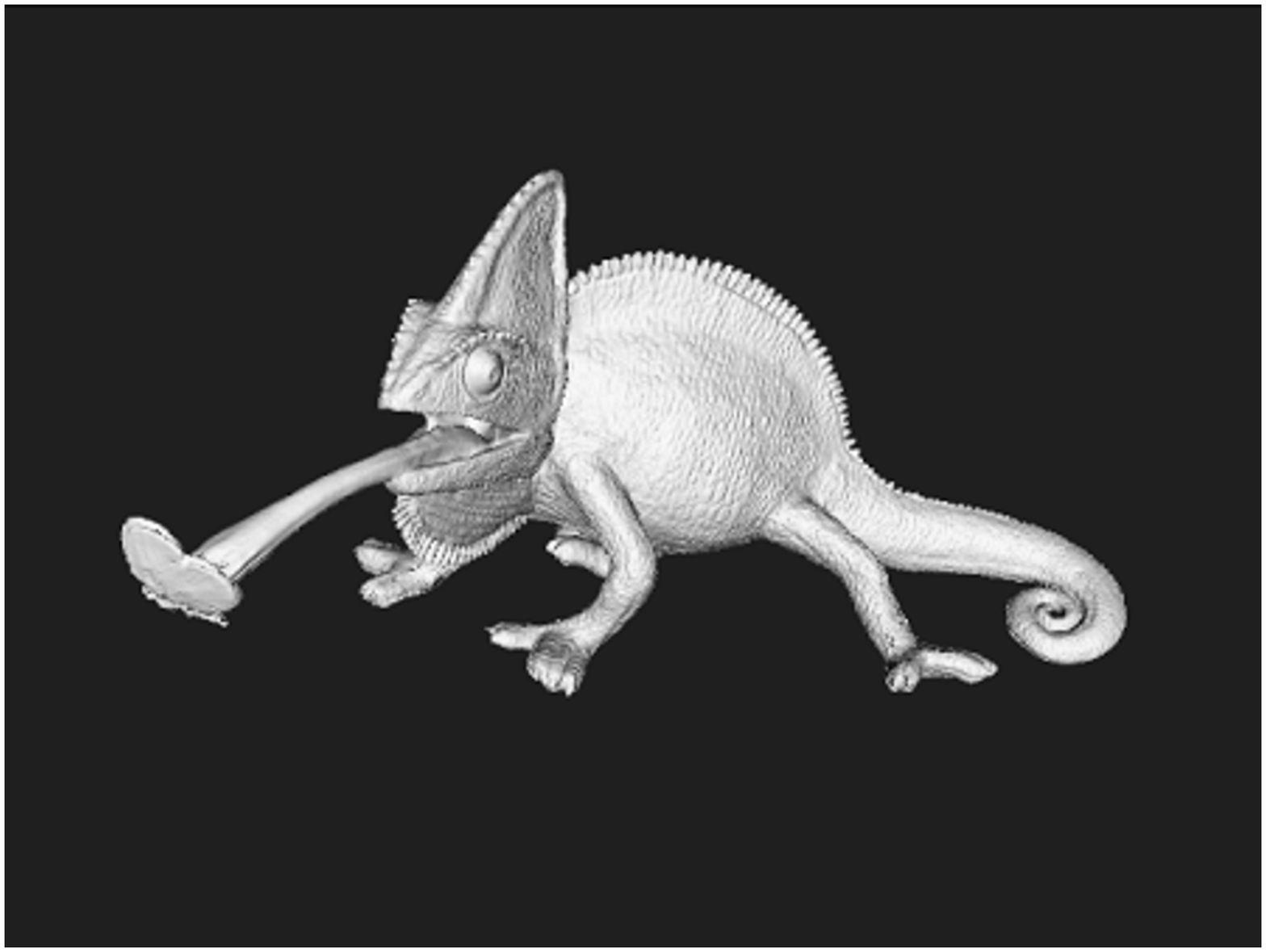}
\subcaption{chameleon}
\label{Fig:LabModel_Chameleon}
\end{minipage}
\begin{minipage}{.19\textwidth}
\includegraphics[width=\textwidth]{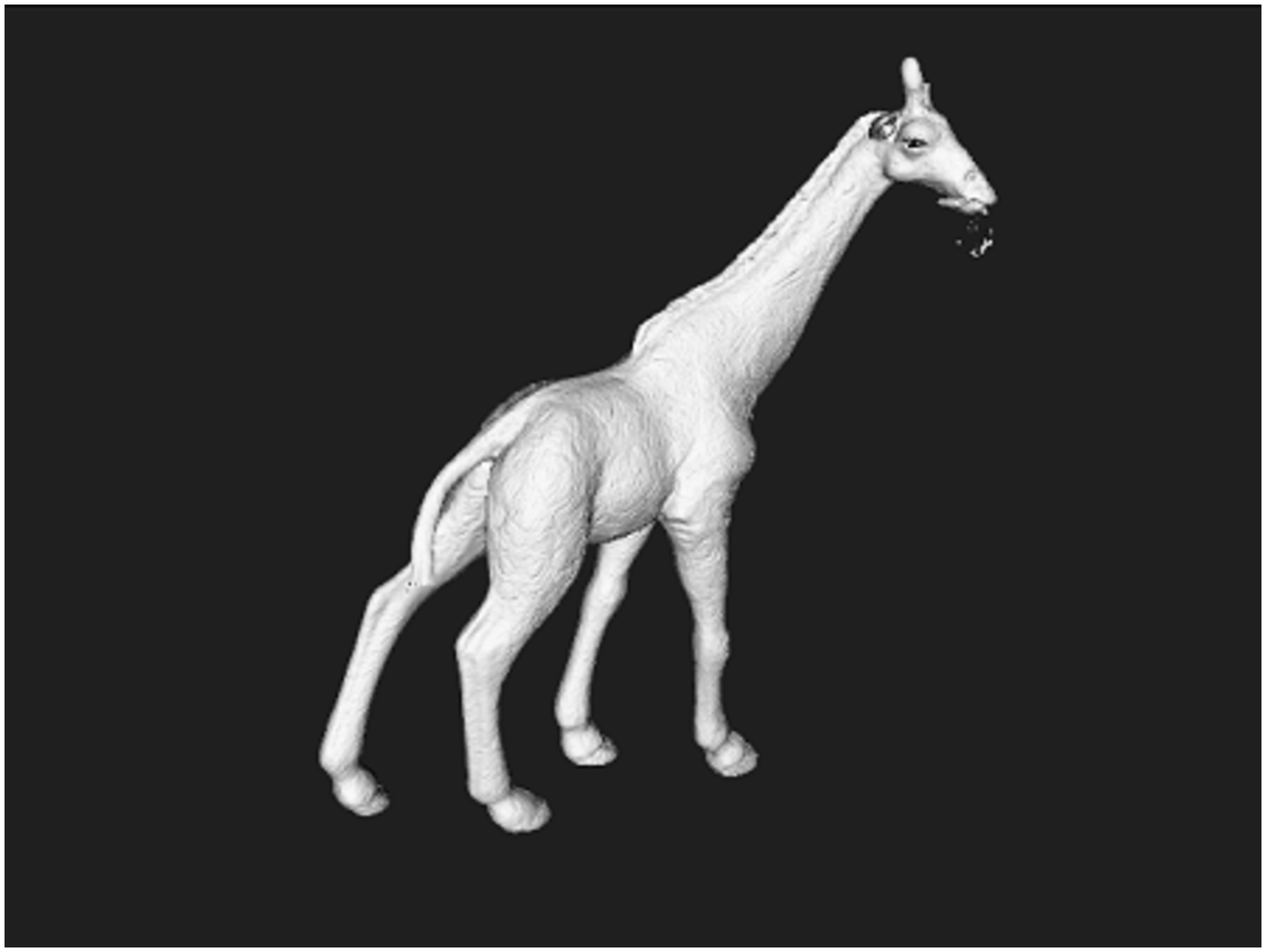}
\subcaption{giraffe}
\label{Fig:LabModel_Giraffe}
\end{minipage}

\begin{minipage}{.19\textwidth}
\includegraphics[width=\textwidth]{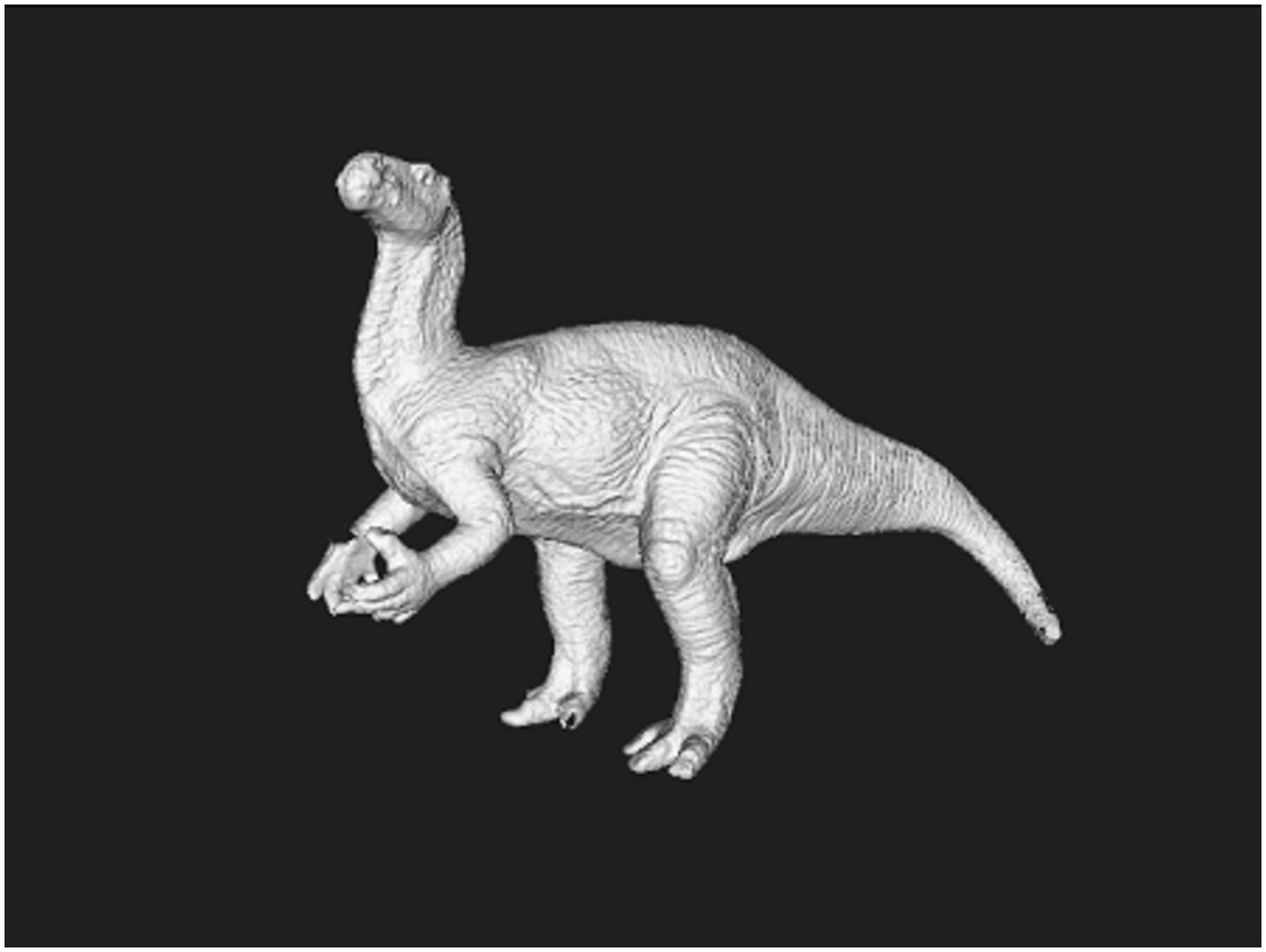}
\subcaption{dinosaur}
\label{Fig:LabModel_Dinosaur}
\end{minipage}
\begin{minipage}{.19\textwidth}
\includegraphics[width=\textwidth]{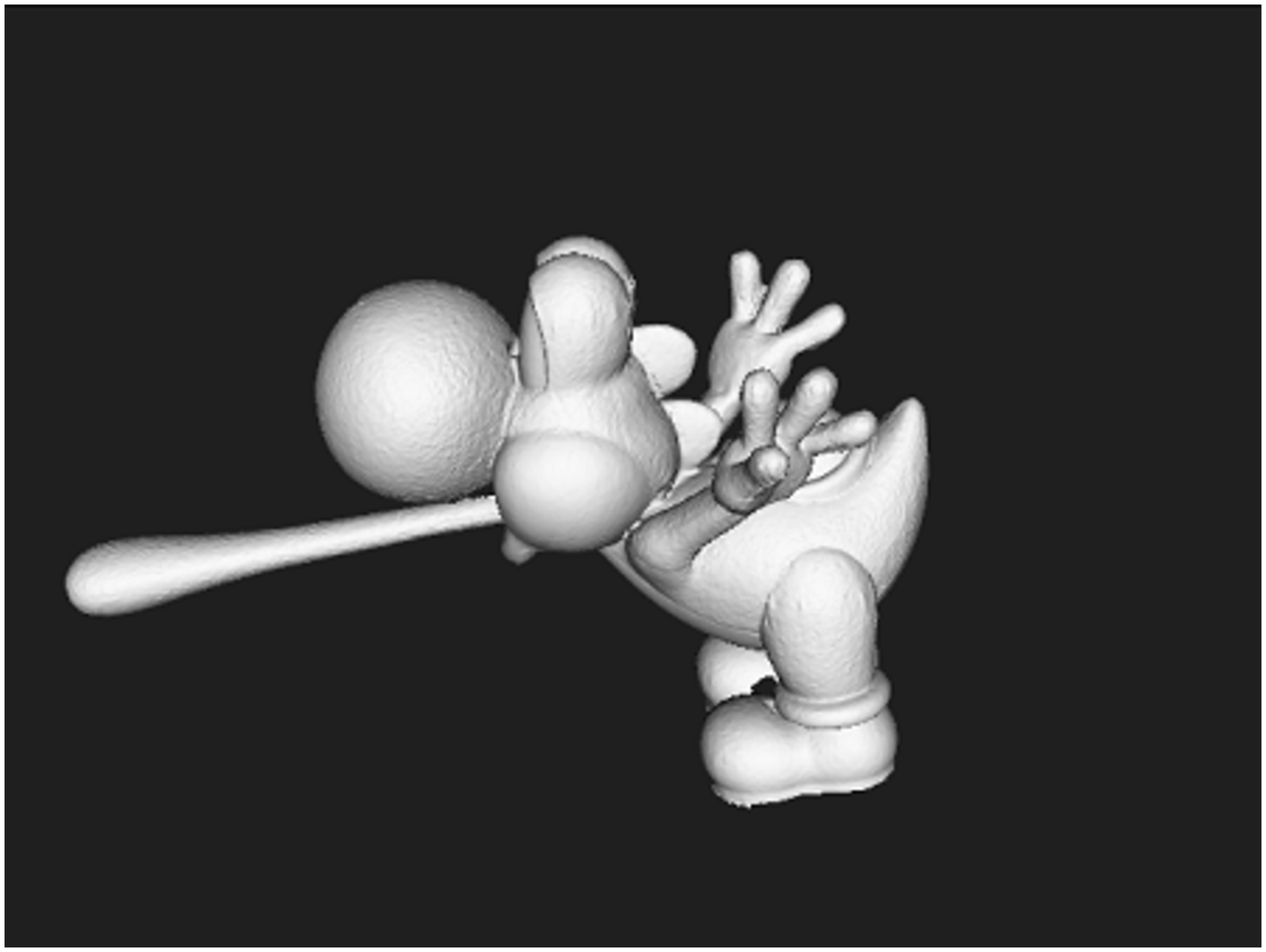}
\subcaption{Yoshi}
\label{Fig:LabModel_Yoshi}
\end{minipage}
\begin{minipage}{.19\textwidth}
\includegraphics[width=\textwidth]{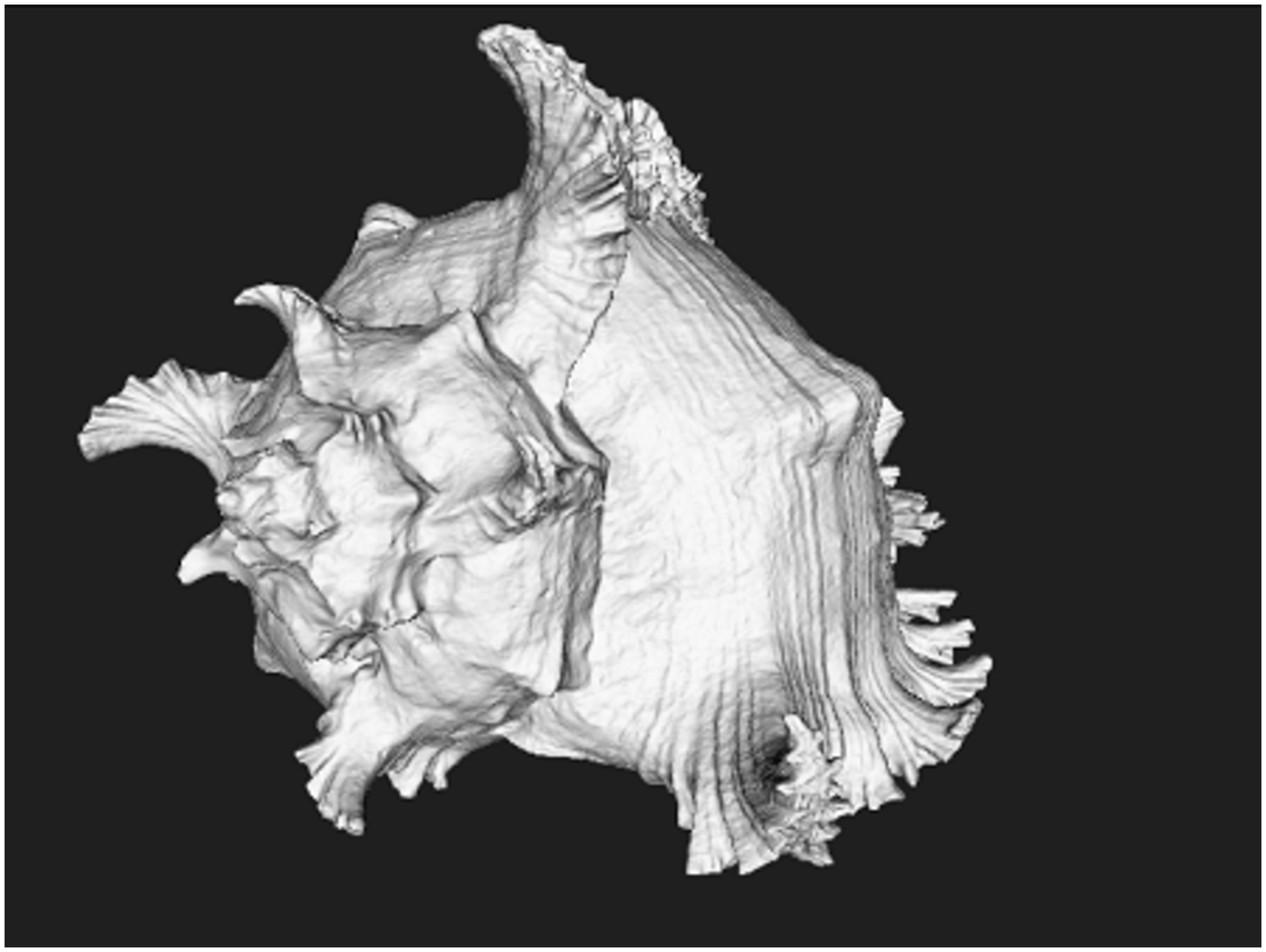}
\subcaption{shell}
\label{Fig:LabModel_Shell}
\end{minipage}
\begin{minipage}{.19\textwidth}
\includegraphics[width=\textwidth]{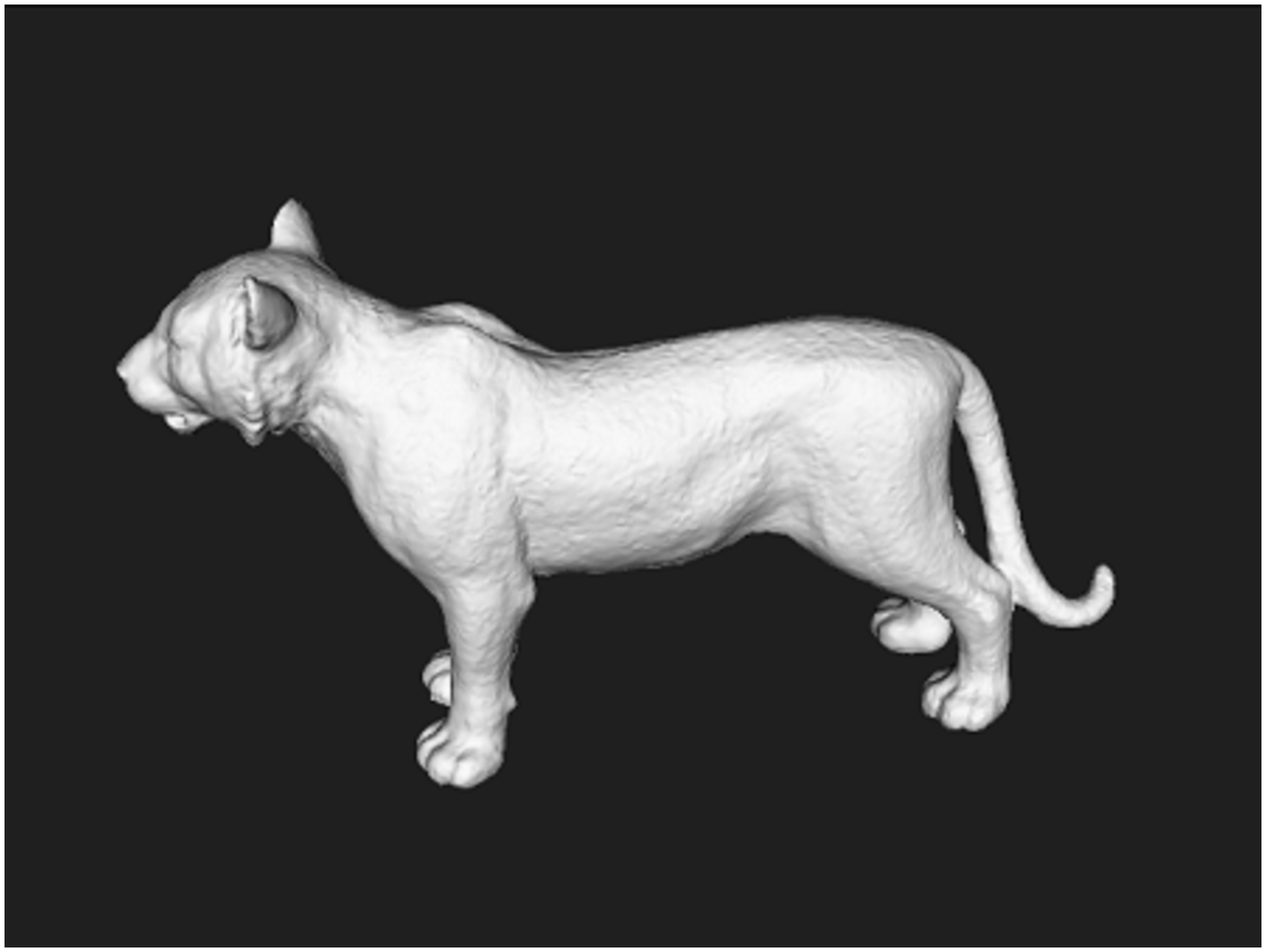}
\subcaption{tiger}
\label{Fig:LabModel_Tiger}
\end{minipage}

\caption{Laboratory data}
\label{fig:labModel}
\end{figure*}

For evaluation, we used partial point cloud data converted from PLY data
of Stanford University\cite{StanfordPly} and from PLY data measured in
the laboratory. Each image is like
fig.~\ref{fig:stanfordModel}, \ref{fig:labModel}. Each partial point
cloud data was obtained by rotating the model by 20 degrees about the
Y-axis direction, so there were 18 ring-shaped point cloud data for each
model. Evaluation was performed between adjacent data. There are 5
models of Stanford and 9 models of laboratory, so the total number of
registration trials is \((5 + 9) \times 18 = 252\).

\subsection{Way of evaluation}

In this experiment, Root Mean Square Error (RMSE) is used as the
evaluation method on accuracy. Since the ground truth of the rigid transformation is
known for the data used in this experiment, the evaluation value is
calculated by the following equation using the error between the
estimated value and the ground truth.

\begin{equation}
\begin{aligned}
E_{\rm RMSE} =& \sqrt{\frac{E^2_X + E^2_Y}{|X|+|Y|}} \\
& E^2_X = \sum_{{\bf x}\in X} \left\| \tilde{\mathcal{T}}^{-1}{\bf x} - {\bf x} \right\|^2 \\
& E^2_Y = \sum_{{\bf y}\in Y} \left\| \tilde{\mathcal{T}}{\bf y} - {\bf y} \right\|^2 
\end{aligned}
\end{equation}

Where the two point clouds used for registration are
\(X,~Y \) \(\subset \{ (x~y~z~1)^{\rm T} \) \(|~x,~y,~z \in \mathbb{R} \}\), and
the error matrix is
\(\tilde{\mathcal{T} } = \mathcal{T}_{\rm gt}^{-1} \mathcal{T}\) when
the estimated value of the rigid transformation matrix is
\(\mathcal{T}\) and the ground truth is \(\mathcal{T}_{\rm gt}\).

\subsection{Experimental result}

\begin{figure}
\centering
\includegraphics[width=.9\linewidth]{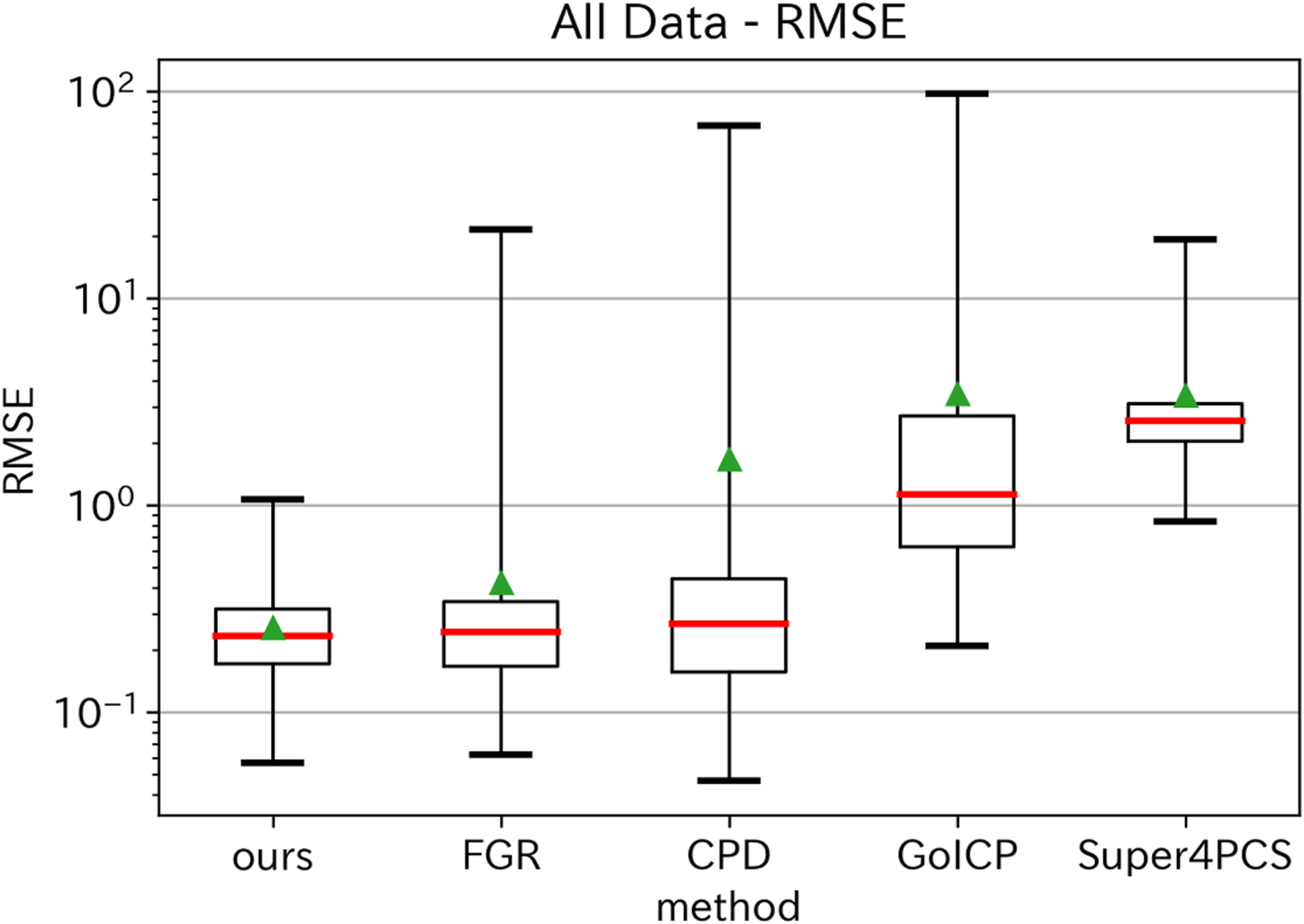}
\caption{RMSE result of experiments (All Data)}\label{fig:RMSE_all}
\end{figure}

\begin{figure}
\centering
\includegraphics[width=.9\linewidth]{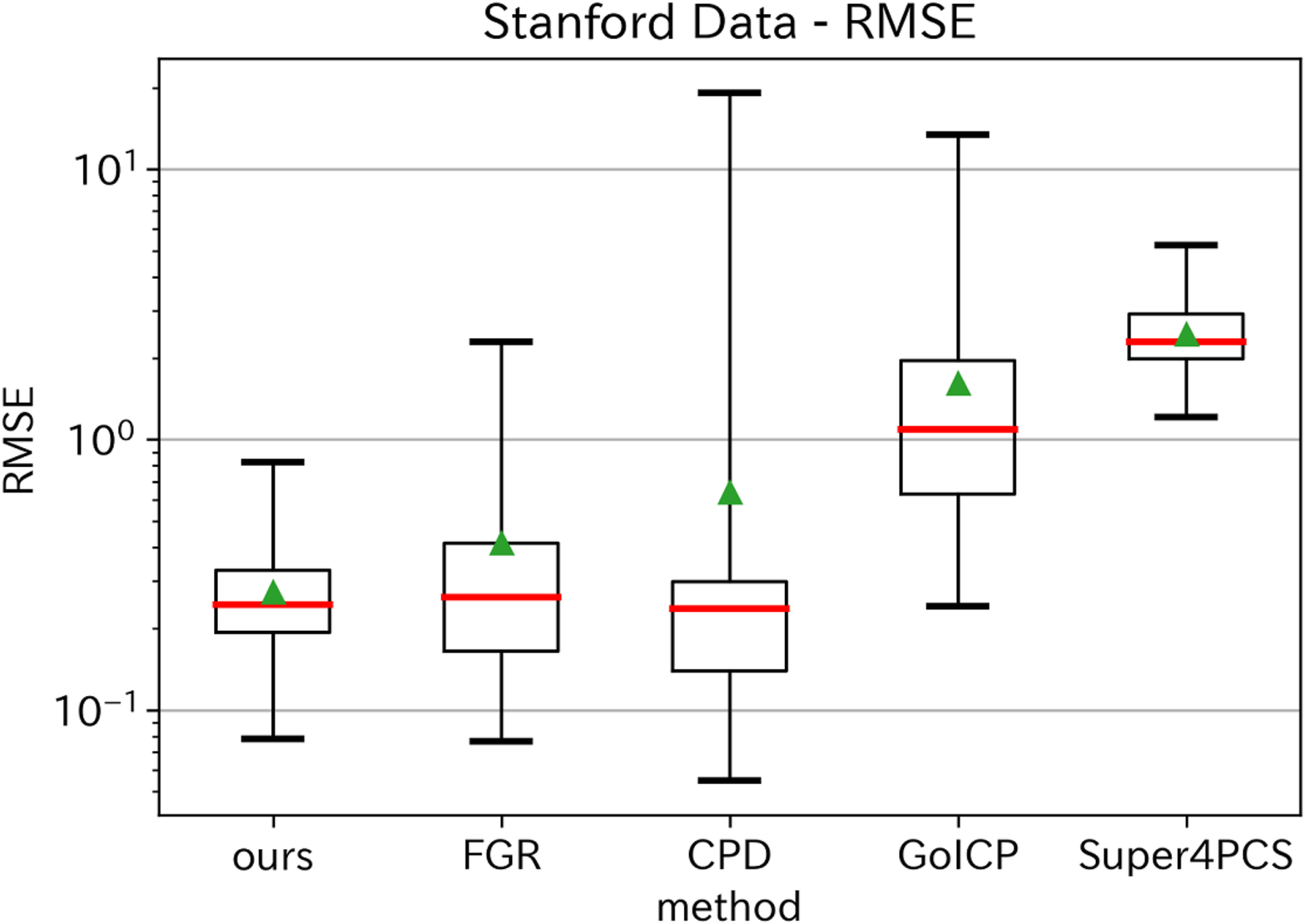}
\caption{RMSE result of experiments (Stanford Data)}\label{fig:RMSE_stanford}
\end{figure}

\begin{figure}
\centering
\includegraphics[width=.9\linewidth]{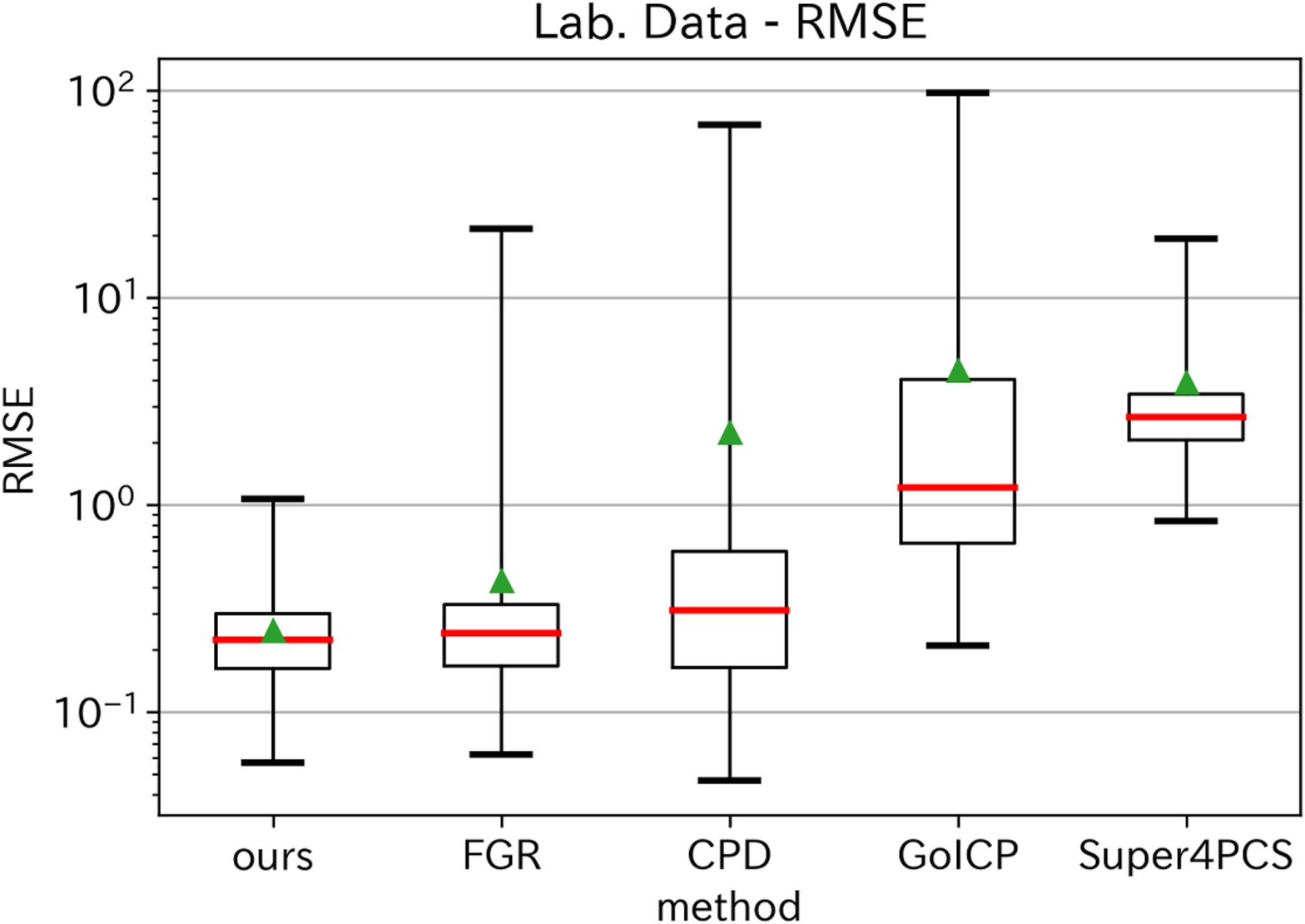}
\caption{RMSE result of experiments (Lab. Data)}\label{fig:RMSE_lab}
\end{figure}

When we conducted experiments on our method and the
comparatives, the overall RMSE results became fig.~\ref{fig:RMSE_all}.
The RMSE results for Stanford's data were fig.~\ref{fig:RMSE_stanford},
and the ones for laboratory data were fig.~\ref{fig:RMSE_lab}. The
results show that our method was the most accurate and stable
for both data. Also, in the comparative methods, the accuracy of the
maximum value was worse in the laboratory data than in the Stanford
data. On the other hand, it can be seen that our method
maintained accuracy.

\begin{figure}
\centering
\includegraphics[width=.9\linewidth]{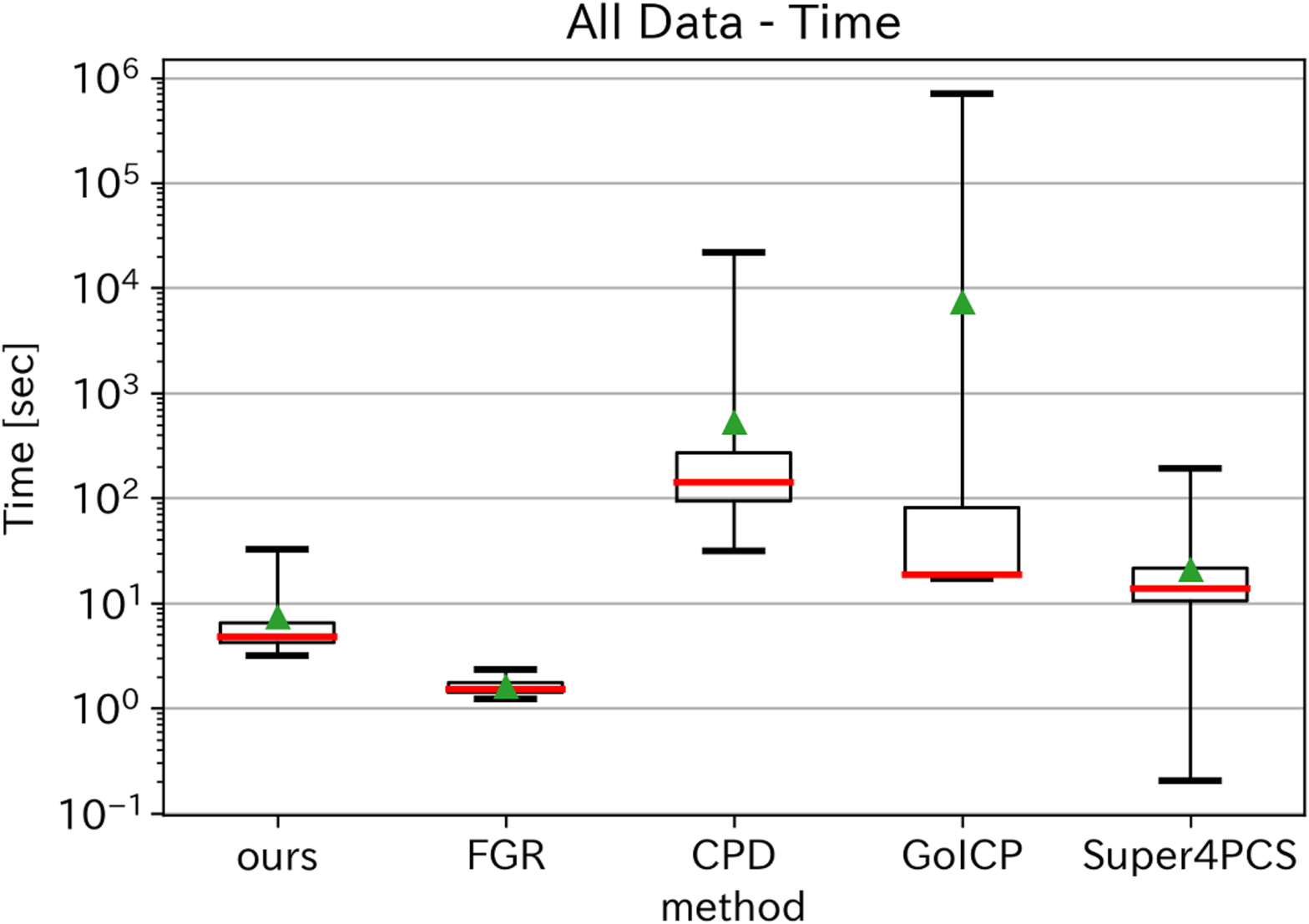}
\caption{Time result of experiments}\label{fig:Time_all}
\end{figure}

The results of actual runtimes of our method and the comparative methods were fig.~\ref{fig:Time_all}.
From the results, we were able to obtain the second highest speed after FGR in terms of median and mean values.

We consider why our method obtained stable results with high
accuracy. First, if the methods worked well, the accuracy of proposed
method did not change significantly from FGR and CPD. This can be seen
from the fact that the medians of fig.~\ref{fig:RMSE_all} are close to
each other. However, with FGR and CPD, there were cases where some
registrations failed. This led to a great lack of stability. Stanford's
Buddha data was the worst, and was significantly inferior to our method in all registrations. The accuracy of the other data
often deteriorated at specific registrations.

The reason for this is that CPD uses all points of point clouds and FGR chooses correspondence, but there is a problem in how to choose.
Similar to our method, in FGR, the correspondences are selected using the FPFH descriptors and constrained by forming triplets.
However, unlike our method, the search of correspondences is performed for all points and this search is stopped when the necessary number has been collected.
Therefore, in some results, FGR selected inappropriate points and did not selected good ones. 
On the other hand, our method excluded inappropriate points by determining the keypoints in advance.
As a result, it is considered that our method was very stable compared to the comparative methods.

In addition, our method could be calculated at the speed next to FGR.
Considering this reason,
first, by using keypoints, our method reduces the point cloud.
This made it possible to reduce the number of simple evaluations in subsequent calculations.
In addition, although the calculation of triplets is inherently expensive, 
the appropriate one-to-one correspondences are preferentially used by the preliminary evaluation, 
and the calculation of triplets is stopped in the middle.
Furthermore, since our method does not perform iterative calculation such as continuing the evaluation until the result becomes stable, 
the calculation time does not increase due to the unstable result.
As a result, sufficient speed could be obtained.
However, we also found that the calculation of triplets was still expensive, and the algorithm procedure was complicated compared to FGR, so it was not as fast as FGR.

\section{Conclusion}

In this paper, we proposed a global registration method between 3-dimensional point clouds with high accuracy and stability at a practical speed.

In our method, first the keypoints were extracted from the point cloud, and the one-to-one correspondences were
generated using the FPFH descriptors. Furthermore, after obtaining the search priority, we
generated triplets using PPF descriptors. The parameters of the rigid
transformation were calculated from each triplet, and the rotation
vector and translation vector were obtained. Then, histograms were
generated for these rotation and translation vectors, and the mode was
defined as the rigid transformation between the point clouds.

In the experiment, partial point clouds created from the data of Stanford and the laboratory were used.
When compared with other methods such as Fast Global Registration (FGR), our method was sufficiently accurate and had no major failures. 
As a result, our method obtained high accurate and the most stable results.
We think that the reason for this is the selection of keypoints.
FGR may select all points.
Therefore, for some registrations, FGR selected points where the accuracy of the descriptor was unstable.
Other comparative methods likewise failed registration because they used improper points for matching.
On the other hand, our method excluded inappropriate points by determining the keypoints in advance.
As a result, it is considered that our method was very stable compared to the comparative methods.

In addition, we obtained the result that the calculation speed was second only to FGR
 by using keypoints and evaluating the priorities of the one-to-one correspondences before the calculation of triplets.
Although not as fast as FGR,
 our method was fast enough given the complexity of the algorithm.

\section*{Acknowledgement}
\addcontentsline{toc}{section}{Acknowledgement}

We are deeply grateful to everyone in the laboratory. In particular,
Masamichi Kitagawa gives insightful comments and suggestions.

\bibliography{yoshii_d_matome_e}

\begin{thebibliography}{10}

\bibitem{ICPpoint2point1992}
{P. Besl and N. D. McKay}.
\newblock ``A Method for Registration of 3-D Shapes''.
\newblock {\em IEEE Transactions on Pattern Analysis and Machine
  Intelligence(T-PAMI)}, Vol.~14, No.~2, pp. 239--256, 1992.

\bibitem{ICPpoint2plane1991}
{Y. Chen and G. Medioni}.
\newblock ``Object Modeling by Registration of Multiple Range Images''.
\newblock {\em IEEE International Conference Robotics and Automation}, Vol.~3,
  pp. 2724--2729, 1991.

\bibitem{colorICP1999}
{A. E. Johnson and B. K. Sing}.
\newblock ``Registration and integration of textured 3D data''.
\newblock {\em Image and Vision Computing}, Vol.~17, No.~2, pp. 135--147, 1999.

\bibitem{ICPIF2000}
{G. C. Sharp, S. W. Lee and D. K. Wehe}.
\newblock ``ICP Registration using Invariant Features''.
\newblock University of Michigan, Sogang University, 2000.

\bibitem{EM-ICP2002}
{S. Granger and X. Pennec}.
\newblock ``Multi-scale EM-ICP: A fast and robust approach for surface
  registration''.
\newblock {\em European Conference on Computer Vision (ECCV)}, pp. 418--432,
  2002.

\bibitem{LM-ICP2003}
{A. Fitzgibbon}.
\newblock ``Robust registration of 2D and 3D point sets''.
\newblock {\em Image and Vision Computing}, Vol.~21, No.~13, pp. 1145--1153,
  2003.

\bibitem{Go-ICP2013}
{J. Yang, H. Li and Y. Jia}.
\newblock ``Go-ICP: Solving 3D registration efficiently and globally
  optimally''.
\newblock {\em International Conference on Computer Vision (ICCV)}, pp.
  1457--1464, 2013.

\bibitem{Go-ICP2016}
{J. Yang, H. Li, D. Campbell and Y. Jia}.
\newblock ``Go-ICP: A Globally Optimal Solution to 3D ICP Point-Set
  Registration''.
\newblock {\em IEEE Transactions on Pattern Analysis and Machine
  Intelligence(T-PAMI)}, Vol.~38, pp. 2241--2254, 2016.

\bibitem{RegGMM2005}
{B. Jian and B. C. Vemuri}.
\newblock ``A Robust Algorithm for Point Set Registration Using Mixture of
  Gaussians''.
\newblock {\em International Conference on Computer Vision (ICCV)}, Vol.~2, pp.
  1246--1251, 2005.

\bibitem{RegGMMframework2000}
{H. Chui and A. Rangarajan}.
\newblock ``A feature registration framework using mixture models''.
\newblock {\em IEEE Workshop on Mathematical Methods in Biomedical Image
  Analysis}, pp. 190--197, 2000.

\bibitem{GMM_CPD2010}
{A. Myronenko and X. Song}.
\newblock ``Point Set Registration: Coherent Point Drift''.
\newblock {\em IEEE Transactions on Pattern Analysis and Machine
  Intelligence(T-PAMI)}, Vol.~32, No.~12, pp. 2262--2275, 2010.

\bibitem{GMMerge2015}
{D. Campbell and L. Petersson}.
\newblock ``An Adaptive Data Representation for Robust Point-Set Registration
  and Merging''.
\newblock {\em International Conference on Computer Vision (ICCV)}, 2015.

\bibitem{adaptiveGMM2017}
{C. Pu, N. Li and R. B. Fisher}.
\newblock ``Robust Rigid Point Registration based on Convolution of Adaptive
  Gaussian Mixture Models''.
\newblock Cornell University Library, 2017.

\bibitem{4PCS2008}
D.~Aiger, N.~J. Mitra, and D.~Cohen-Or.
\newblock ``4pointss Congruent Sets for Robust Pairwise Surface Registration''.
\newblock {\em ACM Trans. Graph.}, Vol.~27, No.~3, pp. 85:1--85:10, August
  2008.

\bibitem{FGR2016}
Q.-Y. Zhou, J.~Park, and V.~Koltun.
\newblock ``Fast Global Registration''.
\newblock In {\em European Conference on Computer Vision}, 2016.

\bibitem{DO2017}
J.~Vongkulbhisal, F.~D.~l. Torre, and J.~P. Costeira.
\newblock ``Discriminative Optimization: Theory and Applications to Computer
  Vision Problems''.
\newblock {\em The IEEE Conference on Computer Vision and Pattern Recognition
  (CVPR)}, 2017.

\bibitem{SUPER4PCS2014}
N.~Mellado and D.~Mitra, Niloy J.and~Aiger.
\newblock ``Super 4PCS Fast Global Pointcloud Registration via Smart
  Indexing''.
\newblock {\em Computer Graphics Forum}, Vol.~33, No.~5, pp. 205--215, 2014.

\bibitem{ICDO2018}
J.~Vongkulbhisal, B.~Irastorza~Ugalde, F.~D.~l. Torre, and J.~P. Costeira.
\newblock ``Inverse Composition Discriminative Optimization for Point Cloud
  Registration''.
\newblock {\em The IEEE Conference on Computer Vision and Pattern Recognition
  (CVPR)}, pp. 2993--3001, 2018.

\bibitem{FPFH2009_1}
R.~B. Rusu, N.~Blodow, and M.~Beetz.
\newblock ``Fast Point Feature Histograms (FPFH) for 3D Registration''.
\newblock In {\em Proceedings of the 2009 IEEE International Conference on
  Robotics and Automation}, ICRA'09, pp. 1848--1853, Piscataway, NJ, USA, 2009.
  IEEE Press.

\bibitem{FPFH2009_2}
R.~B. Rusu, A.~Holzbach, N.~Blodow, and M.~Beetz.
\newblock ``Fast Geometric Point Labeling Using Conditional Random Fields''.
\newblock In {\em Proceedings of the 2009 IEEE/RSJ International Conference on
  Intelligent Robots and Systems}, IROS'09, pp. 7--12, Piscataway, NJ, USA,
  2009. IEEE Press.

\bibitem{PPF2010}
B.~Drost, M.~Ulrich, N.~Navab, and S.~Ilic.
\newblock ``Model globally, match locally: Efficient and robust 3D object
  recognition''.
\newblock In {\em 2010 IEEE Computer Society Conference on Computer Vision and
  Pattern Recognition}, pp. 998--1005, June 2010.

\bibitem{EstimateTransformation1991}
{S. Ueyama}.
\newblock ``Least-Square Estimation of Transformation Parameters Between Two
  Point Patterns''.
\newblock {\em IEEE Transactions on Pattern Analysis and Machine
  Intelligence(T-PAMI)}, Vol.~13, No.~4, pp. 376--380, 1991.

\bibitem{Freedman1981}
D.~Freedman and P.~Diaconis.
\newblock ``On the histogram as a density estimator:L2 theory''.
\newblock {\em Zeitschrift f{\"u}r Wahrscheinlichkeitstheorie und Verwandte
  Gebiete}, Vol.~57, No.~4, pp. 453--476, Dec 1981.

\bibitem{Rusu2013}
R.~B. Rusu.
\newblock {\em Semantic 3D Object Maps for Everyday Robot Manipulation}.
\newblock Springer Publishing Company, Incorporated, 2013.

\bibitem{StanfordPly}
``{\mbox{The Stanford 3D Scanning Repository}}''.
\newblock {http://graphics.stanford.edu/data/3Dscanrep/}.

\end{thebibliography}
\bibliographystyle{junsrt}

\end{document}